\documentclass[preprint,12pt]{elsarticle}

\usepackage{diagbox}

\usepackage{hyperref}
\usepackage{url}
\usepackage{soul}

\usepackage[linesnumbered,boxed,ruled,commentsnumbered]{algorithm2e}
\usepackage[normalem]{ulem}
\useunder{\uline}{\ul}{}
\usepackage{amssymb}
\usepackage{amsmath}
\usepackage{cases}
\usepackage{amstext}
\usepackage{graphicx}
\usepackage{amsthm}
\usepackage{booktabs}
\usepackage{balance}  
\usepackage{xspace,multirow,multicol,colortbl,xcolor,tabularx}
\usepackage{subfig}
\usepackage{subfloat}

\usepackage{comment}
\usepackage{flushend}
\usepackage{enumitem}
\usepackage{pgfplots}
\pgfplotsset{compat=1.16}

\def\header{\vspace{1mm} \noindent}

\newcommand{\ie}{{\it i.e.},\xspace}
\newcommand{\eg}{{\it e.g.},\xspace}

\newcommand{\ours}{SST\xspace}
\newcommand{\oursG}{SSTGCN\xspace}
\newcommand{\oursD}{SSTDA\xspace}

\newtheorem{lemma}{Lemma}

\newtheorem{definition}{Definition}

\newcommand{\graph}{\mathcal{G}\xspace}
\newcommand{\vertice}{\mathcal{V}\xspace}
\newcommand{\feature}{\mathbf{X}\xspace}
\newcommand{\edge}{\mathcal{E}\xspace}

\newcommand{\labelset}{\mathcal{L}\xspace}
\newcommand{\unlabelset}{\mathcal{U}\xspace}
\newcommand{\classifier}{f\xspace}

\newcommand{\loss}{L\xspace}
\newcommand{\losspop}{L_{pop}\xspace}
\newcommand{\losspse}{L_{pse}\xspace}
\newcommand{\losssst}{L_{sp}\xspace}

\newcommand{\real}{\mathbb{R}\xspace}

\newcommand{\classes}{\mathcal{C}\xspace}

\newcommand{\vs}{{\it v.s.},\xspace}

\newcommand{\EE}{\mathbb{E}\xspace}
\newcommand{\PP}{\mathbf{P}\xspace}
\newcommand{\norm}[1]{\left\|#1\right\|\xspace}

\newcommand{\numOfVertice}{n\xspace}
\newcommand{\numOfClasses}{c\xspace}
\newcommand{\dimFeatures}{d\xspace}

\newcommand{\bX}{\mathbf{X}\xspace}

\newcommand{\by}{\mathbf{Y}\xspace}
\newcommand{\bY}{\mathbf{Y}\xspace}
\newcommand{\bC}{\classes}

\newcommand{\bZ}{\mathbf{Z}\xspace}

\newcommand{\bs}{\mathbf{s}\xspace}
\newcommand{\bS}{\mathbf{S}\xspace}

\newcommand{\argmax}{\mathop{\arg\max}\xspace}

\newcommand{\bH}{\mathbf{H}\xspace}
\newcommand{\bA}{\mathbf{A}\xspace}
\newcommand{\bD}{\mathbf{D}\xspace}
\newcommand{\bF}{\mathbf{F}\xspace}

\newcommand{\bU}{\mathrm{U}\xspace}

\newcommand{\bW}{\mathbf{W}\xspace}

\newcommand{\bI}{\mathbf{I}\xspace}

\definecolor{forestgreen}{RGB}{34, 139, 34}

\definecolor{RYB1}{RGB}{192, 128, 255}
\definecolor{RYB2}{RGB}{255, 192, 32}
\definecolor{RYB3}{RGB}{139, 0, 0}
\definecolor{RYB4}{RGB}{0, 128, 255}

\newenvironment{customlegend}[1][]{%
    \begingroup
    \csname pgfplots@init@cleared@structures\endcsname
    \pgfplotsset{#1}%
}{%
    \csname pgfplots@createlegend\endcsname
    \endgroup
}%

\def\addlegendimage{\csname pgfplots@addlegendimage\endcsname}

\makeatletter
\newcommand\footnoteref[1]{\protected@xdef\@thefnmark{\ref{#1}}\@footnotemark}
\makeatother

\definecolor{mygr}{HTML}{e6e6e6}

\journal{Information Sciences}

\begin{document}

\begin{frontmatter}



\title{Effective Stabilized Self-Training on Few-Labeled Graph Data}


\author[a]{Ziang Zhou}
\ead{20071642r@connect.polyu.hk}

\author[a]{Jieming Shi \corref{cor1}}
\ead{jieming.shi@polyu.edu.hk}
\cortext[cor1]{Corresponding author}

\author[b]{Shengzhong Zhang}
\ead{szzhang17@fudan.edu.cn}

\author[b]{Zengfeng Huang}
\ead{huangzf@fudan.edu.cn}

\author[a]{Qing Li}
\ead{csqli@comp.polyu.edu.hk}

\address[a]{{Department of Computing, The Hong Kong Polytechnic University}, {Hong Kong}, {China}}


\address[a]{{School of Data Science, Fudan University}, {Shanghai}, {China}}

\begin{abstract}
Graph neural networks (GNNs) are designed for semi-supervised node classification on graphs where only a subset of nodes have class labels. However, under extreme cases when very few labels are available (\eg 1 labeled node per class), GNNs suffer from severe performance degradation. Specifically, we observe that existing GNNs suffer from \textit{unstable} training process on few-labeled graphs, resulting to inferior performance on node classification. Therefore, we propose an effective framework, \textit{Stabilized Self-Training (\ours)}, which is applicable to existing GNNs to handle the scarcity of labeled data, and consequently, boost classification accuracy.
We conduct thorough empirical and theoretical analysis to support our findings and motivate the algorithmic designs in \ours.
We apply  \ours to two popular GNN models GCN and DAGNN, to get \oursG and \oursD methods respectively, and evaluate the two methods against 10 competitors over 5 benchmarking datasets. Extensive experiments show that the proposed \ours framework is highly effective, especially when few labeled data are available.
Our methods achieve superior performance under almost all settings over all datasets.  For instance, on a Cora dataset with only 1 labeled node per class, the accuracy of \oursG is $62.5\%$, $17.9\%$ higher than GCN, and the accuracy of \oursD is $66.4\%$, which outperforms DAGNN by $6.6\%$.
\end{abstract}


\begin{highlights}
\item Conduct thorough analysis on node classification over few-labeled graphs
\item Propose a stabilized self-training framework to improve performance 
\item Evaluate the performance of proposed methods on real graph data
\end{highlights}


\end{frontmatter}



\section{Introduction} \label{sec:intro}
A graph models the relationships between objects as the edges between nodes.
Graphs  are ubiquitous with a wide range of real-world applications, \eg 
social network analysis \citep{qiu2018social,li2019encoding}, traffic  prediction \citep{LIANG2023161}, protein interface prediction \citep{fout2017protein}, and recommendation systems \citep{fan2019recommendation,LIAO2022595}. 
An important task to support these applications is node classification that aims to classify the nodes in a graph into various classes. However, effective node classification is challenging, especially due to  the lack of sufficient labeled data that are expensive to obtain.

To mitigate the issue, semi-supervised node classification on graphs has attracted much attention  \citep{liu2020towards,klicpera2018predict}. It leverages a small amount of labeled nodes  as well as the unlabeled nodes in a graph to train an accurate classification model.
There exists a collection of Graph Neural  Networks (GNNs) for semi-supervised node classification \citep{liu2020towards,klicpera2018predict,kipf2016semi,hamilton2017inductive}.
For instance, 
Graph convolution networks (GCNs) rely on a message passing scheme via graph convolution operations to aggregate  the neighborhood information of a node to generate node representation, which can then be used in downstream classification tasks. 
Despite the great success of GCNs, under the extreme cases when very few labels are given (\eg only 1 labeled node per class), the shallow GCN architecture, typically with two layers, cannot effectively propagate the training labels over the input graph, leading to inferior performance as shown in the experiments (Tables \ref{tbl:exp Pubmed Corafull}, \ref{tbl:exp Cora Citeseer} and \ref{tbl:exp ogbn-arxiv}).  
Recently, several studies try to improve classification accuracy by designing deeper GNN architectures, \eg DAGNN \citep{liu2020towards}. However,   deep GNNs are still not directly designed to tackle the scarcity of labeled data.

\begin{figure*}[t]
\centering\subfloat[GCN ]{
  \includegraphics[scale=0.31]{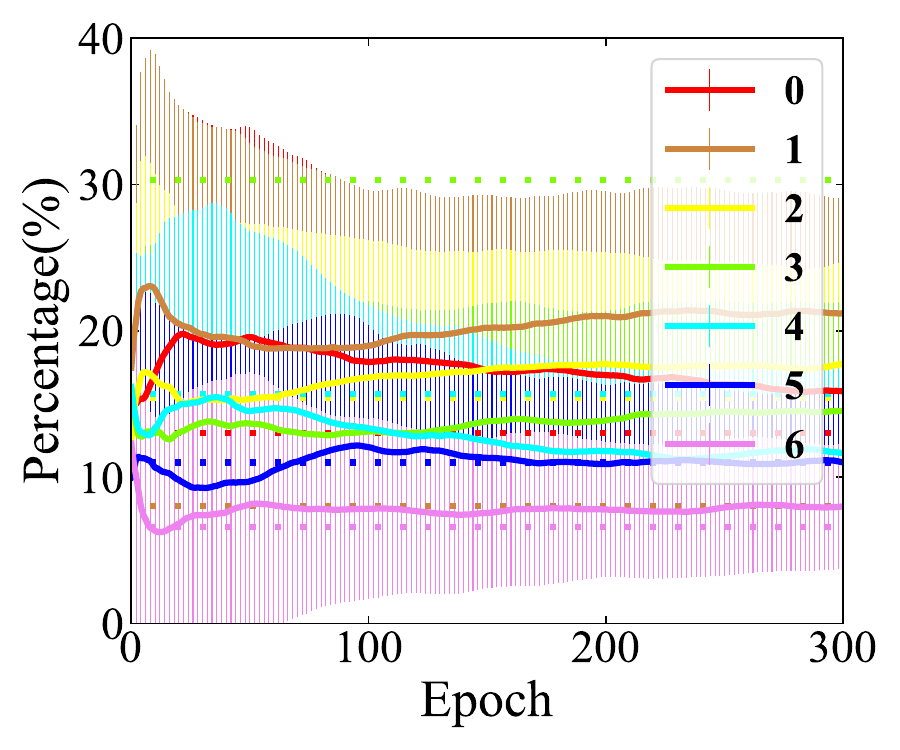}}\label{subfig:dist2}\hspace{6mm}
\centering
\subfloat[DAGNN ]{
  \includegraphics[scale=0.31]{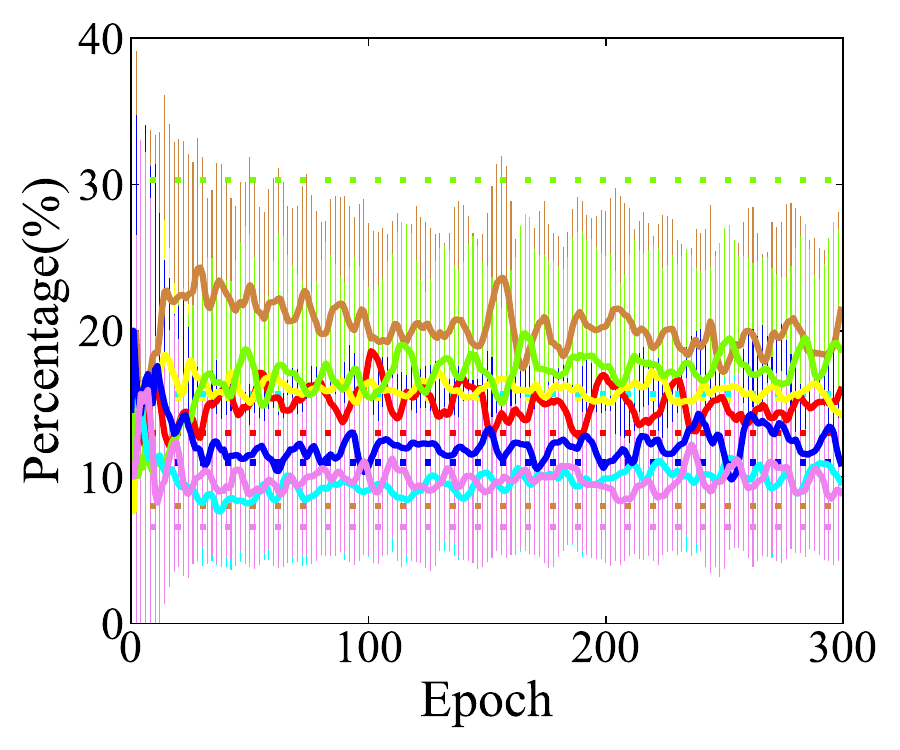}}\label{subfig:dist1}\vspace{0mm}
\vfill
\centering\subfloat[\oursG]{
  \includegraphics[scale=0.31]{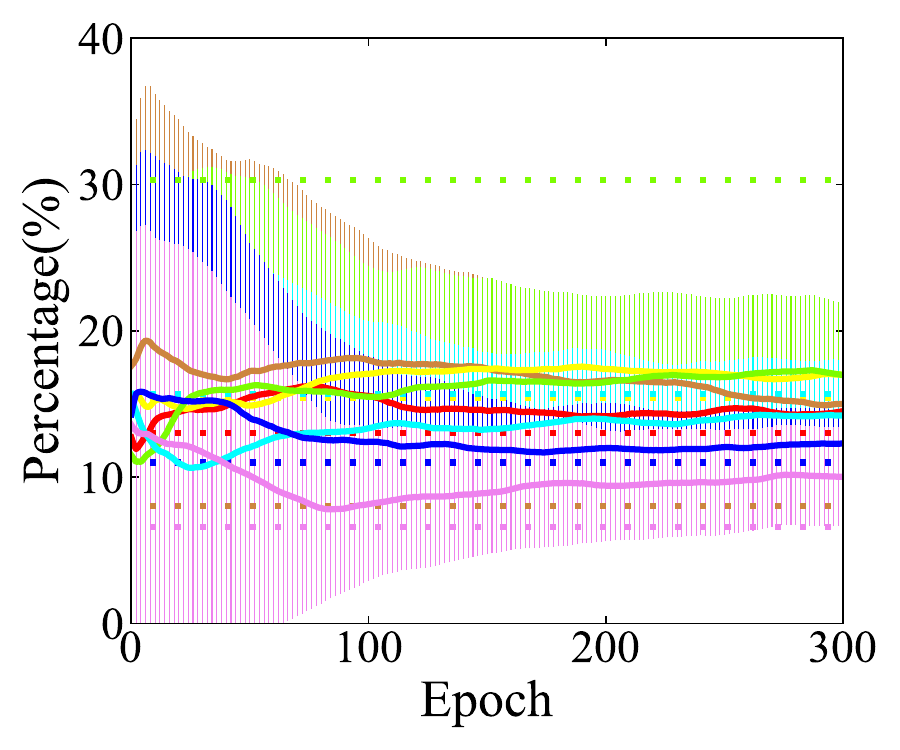}}
\hspace{6mm}
\centering
\subfloat[\oursD]{
  \includegraphics[scale=0.31]{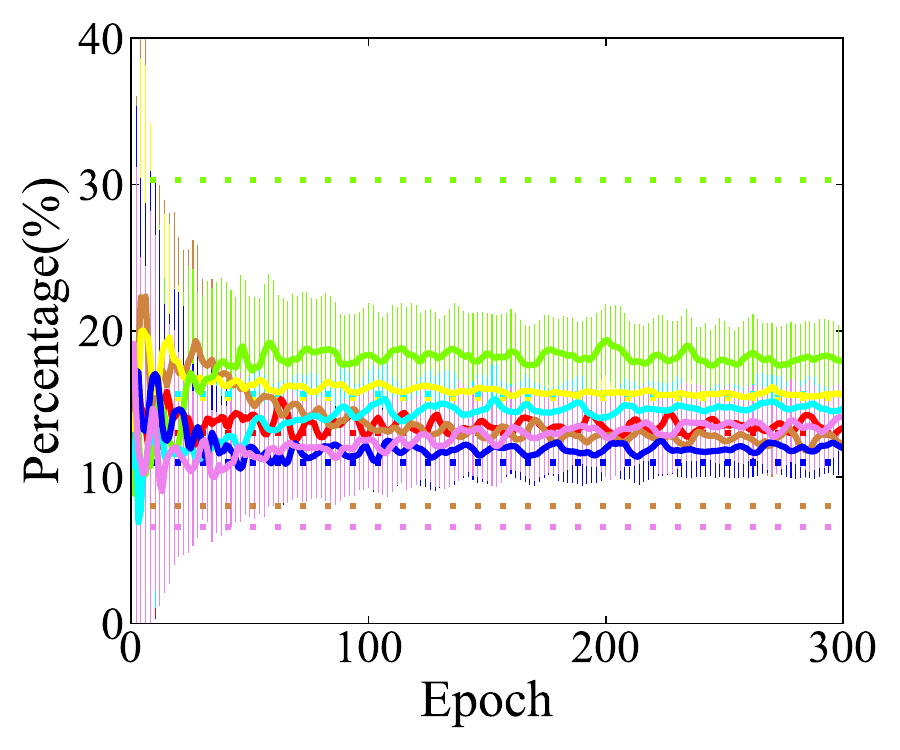}}

\caption{The distribution of predicted labels in different classes in Cora-$1$.}\label{fig:dist}
\end{figure*}

After conducting an in-depth study, we have an interesting finding that existing GNNs suffer from an issue of \textit{unstable} training, when labeled nodes are few.
In particular, on a Cora dataset with 7 classes (see Section \ref{sec:experiments} for dataset details), for each run, we randomly select 1 labeled node per class as the training data (denoted as Cora-1).
Then on Cora-1, we run GCN and DAGNN 100 times with 300 epochs per run, and get the average number of predicted labels per class  with standard deviation  at every epoch  in percentage.
The statistical results of GCN and DAGNN are shown in Figures \ref{fig:dist}(a) and \ref{fig:dist}(b) respectively, where $x$-axis is the epoch from 0 to 300, and $y$-axis is the percentage of a class in the predicted node labels.
There are 7 colored curves representing the average percentage of the predicted labels of the respective classes.
The dashed straight lines are the ground-truth percentage of every class in the Cora dataset.
The shaded areas in color  represent the standard deviation.
Observe that in Figure \ref{fig:dist}(a), GCN has high variance at different runs when predicting node labels, and the variance keeps large at late epochs, \eg 300, which indicates that GCN is quite unstable at different runs with 1 training label per class sampled randomly, leading to inferior classification accuracy as illustrated in our experiments.
As shown in Figure \ref{fig:dist}(b), DAGNN also suffers from the issue of unstable training. The variance of DAGNN is relatively smaller than that of GCN, which provides a hint why DAGNN performs better than GCN.
Nevertheless, both GCN and DAGNN yield unstable training process with large variance.
Since there is only 1 labeled node per class for training, the randomly sampled training nodes can heavily influence the message passing process in GCN and DAGNN, depending on the connectivity of the training nodes to their neighborhoods over the graph topology, which result to the unstable training process observed above. 
In literature, there exists a collection of self-training techniques that enhance the training data by using predicted labels as \emph{pseudo labels} for training \citep{li2018deeper,sun2019multi}. However, as identified in our empirical analysis in Section \ref{sec:empiricalanalysis}, these methods generate imbalanced pseudo labels, which could affect the performance.

To address the above issues when  few labeled data are available, we propose a Stabilized Self-Training (\ours) framework, which is readily applicable to existing GNNs to improve classification accuracy via stabilized training process.
We first conduct thorough empirical and theoretical analysis to identify and explain the issues of existing methods when trained with few labeled nodes for classification. 
Motivated by the analysis, in the proposed \ours framework, we select a set of nodes with predicted labels of high confidence as high-quality pseudo labels, and add such pseudo labels into training data to enhance the training of next epoch, while filtering out the low-confidence predicted labels.  To tackle the unstable training issue of existing GNNs, we develop a stabilized pseudo labeling technique to balance the importance of different classes in training. We then design a negative sampling regularization technique over pseudo labels to further improve node classification accuracy. 
In experiments, we apply our \ours framework to GCN and DAGNN to get methods \oursG and \oursD respectively. 
Figures \ref{fig:dist}(c) and \ref{fig:dist}(d) report the average percentage and standard variance of the predicted labels per class per epoch of \oursG and \oursD on Cora-1 respectively.
Compared with Figure \ref{fig:dist}(a) of GCN, obviously, the variance of \oursG in Figure \ref{fig:dist}(c) decreases quickly and becomes stable as epoch increases.
\oursD is also more stable than DAGNN,  as shown in Figures \ref{fig:dist}(d) and \ref{fig:dist}(b) respectively. 
We conduct extensive experiments on 5 benchmarking datasets, and compare with 10 existing solutions, to evaluate the performance of the proposed \ours framework. 
Experimental results demonstrate that \ours is able to significantly improve classification accuracy of existing GNNs when only few labels are available.
For instance, with the proposed \ours framework, \oursG achieves $62.5\%$ node classification accuracy on Cora-1, significantly improving GCN ($44.6\%$) by $17.9\%$, and \oursD obtains $66.4\%$ accuracy on Cora-1 and outperforms DAGNN ($59.8\%$) by a substantial margin of $6.6\%$.

In summary, our main contributions are as follows:
\begin{itemize}
\item We conduct thorough empirical and theoretical analysis, and make important findings that existing GNNs are unstable when training with few-labeled graph data, and existing self-training techniques suffer from low-quality and imbalanced pseudo labels. Our theoretical analysis provides solid explanations for the findings.
\item We present a Stabilized Self-Training (\ours) framework that can significantly improve classification performance by stabilizing the training process of GNNs. Based on our   analysis, \ours consists of a stabilized pseudo labeling technique and a negative sampling regularization technique over pseudo labels.
\item We apply \ours to popular GNNs, and conduct extensive experiments on 5 datasets to compare against 10 existing methods. The experimental results demonstrate the superior performance of our proposed techniques.
\end{itemize}

The rest of the paper is organized as follows.
We review the related work in Section \ref{sec:relatedwork}. 
In Section \ref{sec:preliminary}, we present the problem formulation of semi-supervised node classification on few-labeled graph data.
We present our analysis and method in Section \ref{sec:analysis}. In particular, we present the empirical analysis in Section \ref{sec:empiricalanalysis}, conduct theoretical analysis in Section \ref{sec:theoreticalanalysis}, and then develop the  \ours framework in Section \ref{sec:algo}.
Extensive experiments are reported in Section \ref{sec:experiments}.
Finally, we conclude the paper in Section \ref{sec:conclusion}.
All proofs are in \ref{secapp:proof}.

\section{Related Work}\label{sec:relatedwork}
In literature, there are two directions to address the scarcity of labeled data for semi-supervised node classification: (i) explore multi-hop graph topological features to propagate the training labels over the input graph \citep{kipf2016semi,liu2020towards}; (ii) enhance the labeled data  by self-training and pseudo labeling \citep{li2018deeper,lee2013pseudo}. Note that these two directions are not mutually exclusive, and they can be applied together on few-labeled graph data. Here we review the existing studies that are most relevant to this paper.

There exist a large collection of GNNs \citep{liu2020towards,klicpera2018predict,kipf2016semi,velickovic2017graph,monti2017geometric,chen2020simple,XU2022799,FU202192}.
We introduce the details of GCN \citep{kipf2016semi} and DAGNN \citep{liu2020towards} here.
  GCN learns the representation of each node by iteratively aggregating the representations of its neighbors. Specifically, GCN consists of $k$ layers, each with the same propagation rule defined as follows. At the $\ell$-th layer, the hidden representations $\bH^{(\ell-1)}$ 
of previous layer are aggregated to get $\bH^{(\ell)}$, 
\begin{equation}\label{equation:propagation rule}
\bH^{(\ell)}=\sigma (\hat{\bA}\bH^{(\ell-1)}\bW^{(\ell)} ),\ell=1,2,...,k.
\end{equation}
$\hat{\bA}=\tilde{\bD}^{-\frac{1}{2}}\tilde{\bA}\tilde{\bD}^{-\frac{1}{2}}$ is the graph laplacian, where $\tilde{\bA}=\bA+\bI$ is the adjacency matrix of $\graph$ after adding self-loops ($\bI$ is the identity matrix) and $\tilde{\bD}$ is a diagonal matrix with $\tilde{\bD}_{ii}=\sum_{j}\tilde{\bA}_{ij}$.   $\bW^{(\ell)}$ is a trainable weight matrix of the $\ell$-th layer, and $\sigma$ is a nonlinear activation function.
Initially, $\bH^{(0)}=\feature$, where $\feature$ is the input feature matrix, describing the node features of all nodes in the input graph. 
Note that GCN usually achieves superior performance with 1 layer or 2 layers. 
When applying many layers to explore large receptive fields, GCN yields degraded performance, due to the over-smoothing issue  \citep{li2018deeper,LEE2022435,chen2020measuring}.
DAGNN addresses the over-smoothing issue  by decoupling  representation transformation and propagation in GNNs \citep{liu2020towards}. Then it utilizes an adaptive adjustment mechanism to balance the information from local and global neighborhoods of every node. Specifically, the mathematical expression of DAGNN is as follows. 
DAGNN uses a learnable parameter $\bs\in\real^{\numOfClasses\times 1}$ to adjust the weight of embeddings at different propagation level (from $1$ to $k$). 
\begin{equation*}
\begin{aligned}
    \bZ=&\text{MLP}(\bX) \in\real^{\numOfVertice\times\numOfClasses} \\
    \bH_{\ell}=& \hat{\bA}^\ell\cdot  \bZ\in\real^{\numOfVertice\times\numOfClasses} ,\ell=1,2,...,k \\
    \bS_{\ell}=& \bH_{\ell}\cdot \bs   \in\real^{\numOfVertice\times 1} ,\ell=1,2,...,k \\
    \hat{\bS}_{\ell}=&  [\bS_{\ell},\bS_{\ell},...,\bS_{\ell}] \in\real^{\numOfVertice\times c} ,\ell=1,2,...,k \\
    \bX_{out}=&\text{softmax}( \sum_{\ell=1}^{k} \bH_{\ell} \odot \hat{\bS}_{\ell} ),
\end{aligned}
\end{equation*}
where $\hat{\bA}^\ell$ is the $\ell$-th power of matrix $\hat{\bA}$, $\odot$ is the Hadamard product, $\cdot$ is dot product, MLP is the Multilayer Perceptron and  softmax operation is on the second dimension.

Apart from GCN and DAGNN, initial GNN studies apply convolution operation in the spectral domain, where the eigenvectors of the graph Laplacian are considered as the Fourier basis \citep{henaff2015deep,defferrard2016convolutional}.
Then GAT \citep{velickovic2017graph} assigns different weights to nodes in the same neighborhood via attention mechanisms.
\citet{monti2017geometric} define convolutions directly in the spatial domain using mixture model CNNs.
APPNP~\citep{klicpera2018predict} adopts a propagation rule based on personalized PageRank \citep{page1999pagerank}, so as to gather both local and global information on graphs.  
GpLCN \cite{FU202192} utilizes the manifold structure information by p-Laplacian matrix to extract abundant features for classification. 
As evaluated in experiments, existing methods yield inferior performance~\citep{liu2020towards,klicpera2018predict,kipf2016semi,velickovic2017graph}, since they are not directly designed to tackle the scarcity of labeled data.

As previously mentioned, another way to address the situation of limited labeled data is to add pseudo labels to training dataset by self-training \citep{li2018deeper}.
Self-training  is a general methodology \citep{scudder1965probability} used in various applications.
\citet{zhou2012self} suggest that selecting informative unlabeled data using a guided search algorithm can significantly improve performance over standard self-training framework. \citet{buchnik2018bootstrapped} mainly consider self-training for diffusion-based techniques.
Recently, self-training has been adopted for semi-supervised tasks on graphs. For instance, \citet{li2018deeper} propose self-training and co-training techniques for GCN. This self-training work selects the top-$k$ confident predicted labels as pseudo labels. 
Co-training technique co-trains a GCN with a random walk model to handle few-labeled data \citep{li2018deeper,chen2011co}.
\citet{sun2019multi} utilize a DeepCluster technique to refine the selected pseudo labels. Compared with existing self-training methods, our framework is different. In particular, our framework has a different strategy to select pseudo labels and also has a stabilizer to address the deficiencies of existing GNNs. Moreover, we propose a negative sampling regularization technique to further boost accuracy. Besides, in existing work, if a node is selected as a pseudo label, it will never be moved out even if the pseudo label becomes obviously wrong in later epochs. On the other hand, in our framework, we update pseudo labels in each epoch to avoid such an issue. 
A recent work \cite{abs-2201-07951} further takes node informativeness
into account for pseudo-label selection.

Further, there are extensive studies on network embedding in recent years, which aims to learn a low-dimensional embedding vector per node  in an unsupervised manner \citep{perozzi2014deepwalk,tang2015line,bojchevski2018deep,velivckovic2018deep}. The learned embedding vectors can be then used in downstream tasks, including node classification.
\citet{perozzi2014deepwalk}  use truncated random walks to learn latent representations, with the assumption that nodes are similar if they are close by random walks. 
\citet{tang2015line} propose to preserve and concatenate the first-order and second-order proximity representations between nodes.
G2G~\citep{bojchevski2018deep} embeds each node as a Gaussian distribution according to a ranking similarity based on the shortest path distances between nodes.
DGI~\citep{velivckovic2018deep} is an embedding method based on GCNs with unsupervised objective that is to maximize mutual information between patch representations and corresponding high-level summaries of graphs. 
However, these unsupervised methods do not leverage labeled data, and thus are not as accurate as our methods in experiments.

\section{Preliminaries}\label{sec:preliminary}

Let $\graph=(\vertice,\edge,\feature)$ be a graph consisting of a node set $\vertice$ with cardinality $n$, a set of edges $\edge$ of size $m$, and a feature matrix $\feature\in \real^{\numOfVertice\times\dimFeatures}$, where $\dimFeatures$ is the number of features. 
An edge in $\edge$ connects two nodes in $\vertice$. 
Every node $v_i\in\vertice$ has a feature vector $\feature_i$ that is the $i$-th row of $\feature$.
Let scalar $c$ be the number of classes, and $\classes$ be the set of class labels ($c=|\classes|$).
We use $\labelset$ to denote the set of labeled nodes, and obviously $\labelset\subseteq\vertice$.
Let $\unlabelset$ be the set of unlabeled nodes and  $\unlabelset=\vertice\setminus\labelset$.
Each labeled node $v_i\in\labelset$ has a one-hot vector $\bY_i\in\{0,1\}^{c}$, indicating the class label of $v_i$.
Under the \emph{few-labeled setting}, $|\labelset|\ll|\unlabelset|$.
The definition of the semi-supervised node classification problem is as follows.

\begin{definition}\label{def:problem}
Given a graph $\graph=(\vertice,\edge,\feature)$, a set of labeled nodes $\labelset\subseteq\vertice$, and a ground-truth class label $\bY_i\in\{0,1\}^{\numOfClasses}$ per node $v_i\in\labelset$, assuming that each node belongs to exactly one class, Semi-Supervised Node Classification predicts the labels of the unlabeled nodes.
\end{definition}

In particular, the aim is to leverage a  graph $\graph$ with the labeled nodes in $\labelset$   to train a     classification model/function $\classifier(\graph,\theta)$.
The model $\classifier$ with trainable parameters $\theta$ outputs a matrix $\bF\in\real^{\numOfVertice\times\numOfClasses}$.
The $i$-th row $\bF_i\in[0,1]^{\numOfClasses}$ represents the   probability vector of node $v_i\in\vertice$. 
We adopt the popular cross-entropy loss to train the model. Given a node $v_i$, its loss $\loss(\bY_i,\bF_i)$ of $\bF_i$ with respect to its true class label $\bY_i$ is 
\begin{equation*}
    \loss(\bY_i,\bF_i)=-\sum_{j=1}^{\numOfClasses}\bY_{i,j} \ln(\bF_{i,j}), 
\end{equation*}
where $\by_{i,j}$ is the $j$-th value in $\by_i$ and $\bF_{i,j}$ is the $j$-th value in $\bF_i$.

During the training process, at a certain epoch, given the matrix $\bF$,
let $\tilde{\by}_{i,j}$ satisfying  Eq. (\ref{eq:pseudolabel}) be the predicted label of node $v_i$ at the epoch. 
In particular, if $\bF_{i,j}$ is the \emph{largest} element in vector $\bF_i$, $\tilde{\bY}_{i,j}$ is 1, otherwise, 0.
We say that, with \emph{confidence} $\bF_{i,j}$, node $v_i$ has class label $\classes_j$. 
\begin{equation}\label{eq:pseudolabel}
    \tilde{\by}_{i,j}=\begin{cases}
     1&\text{if $j=\argmax_{j'}\bF_{i,j'}$},  \\
     0&\text{otherwise},
    \end{cases}
\end{equation}

\section{Our  Method and Analysis}\label{sec:analysis}
We first conduct empirical analysis in Section \ref{sec:empiricalanalysis} to identify the issues of existing techniques, and then perform theoretical analysis  in Section \ref{sec:theoreticalanalysis} to reveal the latent reasons of the empirical findings. 
Lastly, we present our technical designs of Stabilized Self-Training (\ours) in Section \ref{sec:algo}.

\subsection{Empirical Analysis}\label{sec:empiricalanalysis}
We conduct empirical analysis to reveal that existing techniques suffer from imbalanced and low-confidence pseudo labels during the training stage.
We first verify that during the training process, nodes with higher confidence tend to be predicted more accurately. 
On the Cora dataset with 7 classes, each with 1 labeled node for training, Figure \ref{fig:confAccCorr} exhibits the \textit{positive} correlation between the confidence and the accuracy scores of GCN at the 100-th training epoch   with confidence interval 0.1.
In particular, for each confidence interval ($x$-axis), we report the percentage of  unlabeled nodes with predicted labels matching their ground-truth labels (\ie accuracy in $y$-axis). 
Obviously, as the confidence becomes higher, the accuracy increases. On the other hand, if a method generates predictions with low confidence, its performance tends to be inferior.

\begin{figure}[!t]
    \centering
    \subfloat[{\em Confidence \vs accuracy correlation}]{
    \begin{tikzpicture}[scale=0.6]
        \begin{axis}[
            height=\columnwidth*0.48,
            width=\columnwidth*0.7,
            legend style={at={(0.45,1.6)},anchor=north,draw=none,font=\footnotesize,column sep=0.15cm},
            xlabel=\textcolor{black}{\large Intervals of confidence (Epoch 100)}, 
            ylabel=\textcolor{black}{\large {\em Accuracy per Interval (\%)}}, 
            xmin=0, xmax=10,
            ymin=0, ymax=100,
            xtick={0,1,2,3,4,5,6,7,8,9,10},
            xticklabel style = {font=\normalsize,color=black},
            xticklabels={0,0.1,0.2,0.3,0.4,0.5,0.6,0.7,0.8,0.9,1},
            ytick={0,20,40,60,80,100},
            yticklabels={0,20,40,60,80,100},
            scaled y ticks = false,
             yticklabel style={/pgf/number format/fixed,color=black},
            every axis y label/.style={at={(current axis.north west)},right=8mm,above=20mm},
            label style={font=\normalsize,color=black},
            tick label style={font=\normalsize},
            ybar,
            bar width=23pt,
            xtick align=inside,
            nodes near coords,
            nodes near coords align={vertical},
            every node near coord/.append style={font=\normalsize,color=black},
            tick pos=left,
        ]
        \addplot[line width=0.5mm,fill=mygr,draw=black] 
            plot coordinates {
    (0.5,	0)
    (1.5,	25.5)
    (2.5,	36.8)
    (3.5,	45.1)
    (4.5,	58.0)
    (5.5,66.7	)
    (6.5,	56.9)
    (7.5,	75.5)
    (8.5,	65.6)
    (9.5,	89.7)
            };    
        \end{axis}
    \end{tikzpicture}\hspace{8mm}\label{fig:confAccCorr}%
    }%
    \subfloat[{\em Predicted label distribution}]{
    \begin{tikzpicture}[scale=0.6]
        \begin{axis}[
            height=\columnwidth*0.48,
            width=\columnwidth*0.7,
            xlabel=\textcolor{black}{\large Class id (Epoch 20)},
            ylabel=\textcolor{black}{\large {\em Frequency (\%)}},
            xmin=0.5, xmax=7.5,
            ymin=0, ymax=105,
            xtick={1,2,3,4,5,6,7},
            xticklabel style = {font=\normalsize,color=black},
            xticklabels={1,2,3,4,5,6,7},
            ytick={0,20,40,60,80,100},
            yticklabels={0,20,40,60,80,100},
            scaled y ticks = false,
            every axis y label/.style={at={(current axis.north west)},right=3mm,above=20mm},
            label style={font=\normalsize,color=black},
            tick label style={font=\normalsize,color=black},
            ybar,
            bar width=23pt,
            xtick align=inside,
            nodes near coords,
            nodes near coords align={vertical},
            every node near coord/.append style={font=\normalsize,color=black},
            ytick pos=left,
            xtick style={draw=none}
        ]
        \addplot[line width=0.5mm,fill=mygr,draw=black] 
            plot coordinates {
    (1,	    92.4)
    (2,		1.0	)
    (3,		0	)
    (4,     6.6	)
    (5,     0	)
    (6,    0	)
    (7,     0	)
            };    
        \end{axis}
    \end{tikzpicture}\hspace{0.5mm}\label{fig:PseduoImbalance}%
    }%
    
    \caption{Training GCN on Cora with 1 labeled node per class (Cora-1)}
    \vspace{-2mm}
    \end{figure}

 \begin{figure*}[!t]
\centering
\begin{tabular}{ccc}
\hspace{-5mm}

\subfloat[{\em Early epoch: 20}]{
\begin{tikzpicture}[scale=0.58]
    \begin{axis}[
        height=\columnwidth*0.48,
        width=\columnwidth*0.6,
        xlabel={\large Intervals of confidence},
        ylabel={\large\em Frequency (\%)},
        xmin=0, xmax=10,
        ymin=0, ymax=108,
        xtick={0,1,2,3,4,5,6,7,8,9,10},
        xticklabel style = {font=\normalsize,color=black},
        xticklabels={0,0.1,0.2,0.3,0.4,0.5,0.6,0.7,0.8,0.9,1},
        ytick={0,20,40,60,80,100},
        yticklabels={0,20,40,60,80,100},
        scaled y ticks = false,
         yticklabel style={/pgf/number format/fixed},
        every axis y label/.style={at={(current axis.north west)},right=10mm,above=20mm},
        label style={font=\normalsize,color=black},
        tick label style={font=\normalsize,color=black},
        ybar,
        bar width=19pt,
        xtick align=inside,
        nodes near coords,
        nodes near coords align={vertical},
        every node near coord/.append style={font=\footnotesize,color=black},
        tick pos=left,
    ]
    \addplot[line width=0.25mm,fill=mygr,draw=black] 
        plot coordinates {
(0.5,	0)
(1.5,	95.4)
(2.5,	4.6)
(3.5,	0)
(4.5,	0)
(5.5,   0)
(6.5,	0)
(7.5,	0)
(8.5,	0)
(9.5,	0)
        };    
    \end{axis}
\end{tikzpicture}\hspace{0mm}\label{fig:confhist20}%
}%

&

\hspace{-6mm}

\subfloat[{\em Middle epoch:100}]{
\begin{tikzpicture}[scale=0.58]
    \begin{axis}[
        height=\columnwidth*0.48,
        width=\columnwidth*0.6,
        xlabel={\large Intervals of confidence},
        ylabel={\large\em Frequency (\%)},
        xmin=0, xmax=10,
        ymin=0, ymax=108,
        xtick={0,1,2,3,4,5,6,7,8,9,10},
        xticklabel style = {font=\normalsize,color=black},
        xticklabels={0,0.1,0.2,0.3,0.4,0.5,0.6,0.7,0.8,0.9,1},
        ytick={0,20,40,60,80,100},
        yticklabels={0,20,40,60,80,100},
        scaled y ticks = false,
         yticklabel style={/pgf/number format/fixed},
        every axis y label/.style={at={(current axis.north west)},right=10mm,above=20mm},
        label style={font=\normalsize,color=black},
        tick label style={font=\normalsize,color=black},
        ybar,
        bar width=19pt,
        xtick align=inside,
        nodes near coords,
        nodes near coords align={vertical},
        every node near coord/.append style={font=\footnotesize,color=black},
        tick pos=left,
    ]
    \addplot[line width=0.25mm,fill=mygr,draw=black] 
        plot coordinates {
(0.5,	0)
(1.5,	5.1)
(2.5,	34.2)
(3.5,	22.6)
(4.5,	11.9)
(5.5,   8.4)
(6.5,	5.8)
(7.5,	4.9)
(8.5,	3.2)
(9.5,	3.9)
        };    
    \end{axis}
\end{tikzpicture}\hspace{0mm}\label{fig:confhist100}%
}%

&

\hspace{-6mm}

\subfloat[{\em Late epoch: 500}]{
\begin{tikzpicture}[scale=0.58]
    \begin{axis}[
        height=\columnwidth*0.48,
        width=\columnwidth*0.6,
        xlabel={\large Intervals of confidence},
        ylabel={\large\em Frequency (\%)},
        xmin=0, xmax=10,
        ymin=0, ymax=108,
        xtick={0,1,2,3,4,5,6,7,8,9,10},
        xticklabel style = {font=\normalsize,color=black},
        xticklabels={0,0.1,0.2,0.3,0.4,0.5,0.6,0.7,0.8,0.9,1},
        ytick={0,20,40,60,80,100},
        yticklabels={0,20,40,60,80,100},
        scaled y ticks = false,
         yticklabel style={/pgf/number format/fixed},
        every axis y label/.style={at={(current axis.north west)},right=10mm,above=20mm},
        label style={font=\normalsize,color=black},
        tick label style={font=\normalsize,color=black},
        ybar,
        bar width=19pt,
        xtick align=inside,
        nodes near coords,
        nodes near coords align={vertical},
        every node near coord/.append style={font=\footnotesize,color=black},
        tick pos=left,
    ]
    \addplot[line width=0.25mm,fill=mygr,draw=black] 
        plot coordinates {
(0.5,   0)
(1.5,	0.7)
(2.5,	5.2	)
(3.5,	10.6)
(4.5,	10.8)
(5.5,  10.4 )
(6.5,	10.3)
(7.5,	10.1)
(8.5,	13.3)
(9.5,	28.6)
        };    
    \end{axis}
\end{tikzpicture}\hspace{0.5mm}\label{fig:confhist500}%
}%

\end{tabular}
\caption{Confidence distributions of GCN Training on Cora-1.}
\label{fig:confhists}

\end{figure*}

Then in  Figure \ref{fig:confhists}, we provide empirical evidence that most nodes are with low confidence, especially at the early epochs of the training, which will hamper the performance.
Specifically, in Figures \ref{fig:confhist20}, \ref{fig:confhist100}, and \ref{fig:confhist500}, we illustrate the distributions of the confidence scores of all nodes in unlabeled set $\unlabelset$ on Cora, at early (20-th), middle (100-th), and late (500-th) epochs, respectively. Observe that at the early epoch in Figure \ref{fig:confhist20}, $95.4\%$ of the unlabeled nodes are with very low confidence ($<0.2$). If pseudo labels are generated based on the low confidences at early epochs, then such pseudo labels are inaccurate as verified above. Subsequently, they will severely influence the training of later epochs, resulting to unstable training process and unexpected performance degradation.
Further, low-confidence nodes are still  the majority during middle epochs, \eg 100-th epoch in Figure \ref{fig:confhist100}.
The number of low-confidence nodes decreases significantly at very late epochs, \eg 500-th epoch in Figure \ref{fig:confhist500}. 
These observations indicate that the low-confidence nodes keep exposing significant influence during the major part of the whole training process.

Further, Figure \ref{fig:PseduoImbalance} shows the distribution of the predicted labels of the nodes in $\unlabelset$ at the 20-th epoch on Cora in a single run. Obviously, the distribution of the labels is highly imbalanced, \ie $92.4\%$ of the nodes are with class 1 and $6.6\%$ are with class 4. The imbalanced class labels (\ie class 1 in Figure \ref{fig:PseduoImbalance}) will heavily affect the direction of gradient descent during the training process, imposing the model to learn weights resulting to imbalanced predictions, which is often not the ground-truth case. If too many low-confidence nodes are wrongly assigned to a single class at early epochs, existing GNNs, such as GCN, which rely on propagation and transformation over graph topology, will suffer from unstable performance as reported in Figure \ref{fig:dist}.

\subsection{Theoretical Analysis}\label{sec:theoreticalanalysis}
The semi-supervised node classification problem studied in this paper naturally fits the  \textit{transductive learning} setting, since all labeled and unlabeled data in the input graph $\graph$ are known, and no more new data will be added \citep{el2009transductive}. In the following, we analyze the problem from the perspective of \textit{gradient descent}, to theoretically explain the findings made in Section \ref{sec:empiricalanalysis}.

Under the \emph{ideal case} where we have the labels of  \textit{all} nodes in $\vertice$ (the population), given a classifier $\classifier(\graph,\theta)$ with output probability vectors $\bF_i$ for all nodes $v_i$ in $\vertice$, the population loss $\losspop$ is computed as follows. Ideally, the objective is to minimize  $\losspop$ by evaluating \textit{population gradient}  $\nabla_{\theta} \losspop$, and find optimal parameters $\theta^*$. 
\begin{equation*}
    \losspop=\EE_{v_i\sim \bU(\vertice)} \loss(\by_i,\bF_i),
\end{equation*}
where $\bU(\vertice)$ is the uniform distribution over node set $\vertice$.

However, in practice, due to the scarcity of labeled data (\ie $\vert\labelset\vert\ll\vert\vertice\vert$), it is impossible to directly evaluate $\nabla_{\theta} \losspop$. 
Therefore, as explained in Section \ref{sec:relatedwork}, self-training method considers both the labeled nodes and the predicted labels of unlabeled nodes (as pseudo labels) for training.
In particular, during a training epoch, the predicted labels of unlabeled nodes   in Eq. \eqref{eq:pseudolabel} are regarded as the pseudo labels of the nodes. 
Then a \emph{pseudo loss} $\losspse$ is used to approximate the population loss $\losspop$.
Also, \textit{pseudo gradient} $\nabla_{\theta} \losspse$ is used to approximate $\nabla_{\theta} \losspop$, in order to minimize the loss~\citep{gower2019sgd,nesterov2017random}.
Usually the self-training loss $\losspse$ and its gradient $\nabla_{\theta}\losspse$ can be written as 

\begin{align}
    \losspse=&\EE_{v_i\sim \bU(\labelset)}  \loss(\by_i,\bF_i)+\lambda\cdot\EE_{v_i\sim \bU(\unlabelset)}  \loss(\tilde{\by}_i,\bF_i),\label{eq:lossPseudoExpectation} \\
    \nabla_{\theta}\losspse=&\EE_{v_i\sim \bU(\labelset)} \nabla_{\theta} \loss(\by_i,\bF_i)+\lambda\cdot\EE_{v_i\sim \bU(\unlabelset)} \nabla_{\theta} \loss(\tilde{\by}_i,\bF_i),  
    \label{eq:lossPseudoGradient}
\end{align}
where $\bU(\labelset)$ and $\bU(\unlabelset)$ are the uniform distributions over labeled node set $\labelset$ and unlabeled node set $\unlabelset$, respectively.

We rewrite the population gradient $\nabla_{\theta} \losspop$ in a similar format by breaking $\vertice$ into $\labelset$ and $\unlabelset$, 
\begin{equation*}
\begin{aligned}
\nabla_{\theta} \losspop=&\EE_{v_i\sim \bU(\vertice)} \nabla_{\theta}L(\by_i,\bF_i)\\
=&\frac{|\labelset|}{|\vertice|}\EE_{v_i\sim \bU(\labelset)} \nabla_{\theta} \loss(\by_i,\bF_i)+\frac{|\unlabelset|}{|\vertice|}\EE_{v_i\sim \bU(\unlabelset)} \nabla_{\theta} L(\by_i,\bF_i)   .
\end{aligned}
\end{equation*}

Then  we derive a bound on the difference between $\nabla_{\theta} \losspop$ and $\nabla_{\theta} \losspse$ in Eq. \eqref{eq:boundofgradient} with proof  in \ref{proof:boundofgradient}. Specifically, let $\lambda=\frac{|\unlabelset|}{|\labelset|}$, and assume that any gradient satisfies a bounded norm (\ie $\norm{\nabla_{\theta}\loss}\le \Theta$, for any loss $\loss$), which is a common assumption~\citep{zinkevich2010parallelized}. Then the difference between $\nabla_{\theta} \losspop$ and $\nabla_{\theta} \losspse$ is bounded by
\begin{equation}\label{eq:boundofgradient}
\small
    \begin{aligned}
        &\norm{\nabla_{\theta}\losspse-\frac{|\vertice|}{|\labelset|}\nabla_{\theta} \losspop}\\
        =&\frac{|\unlabelset|}{|\labelset|}\cdot\norm{\EE_{v_i\sim \bU(\unlabelset)}\left[ \nabla_{\theta} \loss(\tilde{\by}_i,\bF_i)- \nabla_{\theta} \loss(\bY_i,\bF_i)\right]}\\
        \le & \frac{|\unlabelset|\cdot \PP_{v_i\sim \bU(\unlabelset)}(\tilde{\by}_i\neq \by_i)}{|\labelset|}\cdot\EE_{v_i\sim \bU(\unlabelset)}\left[ \norm{\nabla_{\theta} \loss(\tilde{\by}_i,\bF_i)- \nabla_{\theta} L(\by_i,\bF_i)}\Big| \tilde{\by}_i\neq \by_i \right]  \\
        \le & \frac{2\Theta|\unlabelset|}{|\labelset|}\cdot \PP_{v_i\sim \bU(\unlabelset)}(\tilde{\by}_i\neq \by_i),
    \end{aligned}
\end{equation}
where $\PP_{v_i\sim\bU(\unlabelset)}(\tilde{\by}_i\neq \by_i)$ is the probability that a randomly sampled node $v_i\in\unlabelset$ has a wrongly predicted label.

Obviously, $\PP_{v_i\sim\bU(\unlabelset)}(\tilde{\by}_i\neq \by_i)$ is   the classification error on unlabeled set $\unlabelset$. 
Observe that the bound at the last line in Eq. \eqref{eq:boundofgradient} mainly relies on the quality of pseudo labels $\tilde{\bY}_i$ of nodes $v_i$ in $\unlabelset$. In other words, if we have low-quality pseudo labels from $\unlabelset$, then $\PP_{v_i\sim\bU(\unlabelset)}(\tilde{\by}_i\neq \by_i)$ tends to be large, leading to a large difference between $\nabla_{\theta} \losspop$ and $\nabla_{\theta} \losspse$ and consequently resulting to sub-optimal performance. 
This situation is likely to happen at early epochs when most nodes have low confidence but are still selected as pseudo labels for training as illustrated in Figure \ref{fig:confhist20} of Section \ref{sec:empiricalanalysis}.

Moreover, at the early epochs when many nodes are with low confidence, the imbalanced predicted labels (\eg Figure \ref{fig:PseduoImbalance}) may make $\PP_{v_i\sim\bU(\unlabelset)}(\tilde{\by}_i\neq \by_i)$ even larger.
In the following, we  prove that if we use imbalanced predicted labels as pseudo labels, this will lead to inferior performance. 
To facilitate the analysis, we define that the distribution of ground-truth labels is \textit{$\rho$-balanced} as follows. Specifically, for any two distinct classes $a,b\in\classes$, the balance ratio between them is upper bounded by $\rho$, if and only if
    \begin{equation}\label{eq:balancetruth}
        \begin{aligned}
            \max_{a,b\in\classes}\frac{\PP_{v_i\sim\bU(\unlabelset)}(\by_i=a)}{\PP_{v_i\sim\bU(\unlabelset)}(\by_i=b)}\leq \rho.
        \end{aligned}
    \end{equation}
    
Let $c'=\argmax_{c\in\classes}\PP_{v_i\sim\bU(\unlabelset)}(\tilde{\by}_i=c)$ be  the class with the max predicted probability.
We  define that the predicted labels are \textit{$\eta$-imbalanced} if and only if 
        \begin{equation}\label{eq:unbalancepseudo}
                \max_{b\in\classes,b\neq c'}\frac{\PP_{v_i\sim\bU(\unlabelset)}(\tilde{\by}_i=c')}{\PP_{v_i\sim\bU(\unlabelset)}(\tilde{\by}_i=b)}\geq \eta.
        \end{equation}

Then we present Lemma \ref{lem:error_of_unbalance} that provides a lower bound of the classification error $\PP_{v_i\sim\bU(\unlabelset)}(\tilde{\by}_i\neq \by_i)$.

\begin{lemma}\label{lem:error_of_unbalance}
    If the ground-truth label distribution is $\rho$-balanced and the predicted label distribution is $\eta$-imbalanced, we can get the lower bound of the classification error as follows,
    \begin{equation}
        \PP_{v_i\sim\bU(\unlabelset)}(\tilde{\by}_i\neq \by_i)\geq \frac{\eta}{\eta+|\bC|-1}-\frac{\rho}{\rho+|\bC|-1}.
    \end{equation}
    \end{lemma}

\begin{proof}
    The proof of Lemma \ref{lem:error_of_unbalance} is in \ref{proof:Lemma1}.
\end{proof}

Lemma \ref{lem:error_of_unbalance} indicates that if the predicted labels are highly imbalanced compared with ground-truth labels (\ie $\eta$ is large while $\rho$ is small), the classification error is large.
In practice, the imbalance ratio $\eta$ can be arbitrarily large, particularly at the early training epochs. For instance, as shown in Figure \ref{fig:PseduoImbalance}, the imbalance ratio $\eta$ is up to $14$ when using GCN on Cora. However, the ground-truth balance ratio $\rho$ of Cora is just $4.8$. Based on Lemma \ref{lem:error_of_unbalance}, the large discrepancy between $\eta$ and $\rho$ indicates a large classification error $\PP_{v_i\sim\bU(\unlabelset)}(\tilde{\by}_i\neq \by_i)$, as demonstrated in Figure  \ref{fig:dist}(a) when training with GCN. Further, considering Eq. \eqref{eq:boundofgradient}, this will result in an inaccurate approximation of gradient  $\nabla_{\theta}\losspse$ and lead to ineffective and unstable gradient descent process to train the classification model.

\subsection{Stabilized Self-Training (SST) Framework}\label{sec:algo}
Motivated by the analysis above, in what follows, we develop the SST  framework that not only augments training data with high-quality pseudo labels but also stabilizes the training process, in order to achieve superior performance.

\subsubsection{Stabilized Pseudo Labeling}\label{sec:stableselftraining}

We first explain how to choose pseudo labels and then introduce the loss function of stabilized pseudo labeling.
In particular, in a training epoch, for every unlabeled node $v_i\in\unlabelset$, we first get $N_i$, the number of nodes with the same predicted label as $v_i$, 
\begin{equation}\label{eq:re-weighting}
    N_i=\Big|\Big\{ v_j\in\unlabelset \Big|\tilde{\bY}_{i}=\tilde{\bY}_j \Big\}\Big|
\end{equation}

As shown in Figure \ref{fig:dist}, in existing GCN and DAGNN, the distribution of the predicted labels is unstable with large variance during the training process. We also observe that during the early epochs, most predicted labels are with low confidence (Figure \ref{fig:confhists}), low confidence indicates low accuracy (Figure \ref{fig:confAccCorr}), and the predicted labels are imbalanced (Figure \ref{fig:PseduoImbalance}). If we select too many low-confidence predicted labels as the pseudo labels of the same class, this will hamper the training process and result to inferior performance.
To reduce such unstable situation, we propose to use $N_i$ as a stabilizer for pseudo labeling. 
In particular, in our loss function, we assign weights inversely proportional to $N_i$, to mitigate the imbalance issue. If a class is with a larger $N_i$, its importance will be lower, and vice versa. 
For an unlabeled node $v_i\in\unlabelset$, its predicted label may have low confidence score. We do not want to add  such low-confidence labels into the training of subsequent epochs. Instead, we only choose those unlabeled nodes with high-confidence predicted labels as pseudo labels to be augmented into the training data. In particular, an unlabeled node $v_i$ is selected to be a node with \textit{pseudo label} in next epoch, if its confidence is beyond a threshold $\beta$, as shown below.
We use $\unlabelset'$ to denote all unlabeled nodes selected with pseudo labels,
\begin{equation} \label{eq:betaConfidence}
        \unlabelset'=\Big\{  v_i \in\unlabelset \Big| \max_{j} \bF_{i,j}>\beta  \Big\}
\end{equation}
where $\beta\in[0,1]$ is a threshold controlling the extent of cautious selection for self-training. 
A larger threshold means stricter selection of  pseudo labels.

Then we present the loss of the   stabilized pseudo labeling technique in Eq. (\ref{eq:lossOfOurs}). Specifically, we design $\frac{1}{N_i+1}$ as the stabilizer of the training process, to overcome the deficiencies of existing GNNs illustrated in Figure \ref{fig:dist}. The intuition is that, if an unlabeled node $v_i$ is predicted to be in a class with many pseudo labels, its importance in the loss function should be reduced. In other words, our stabilized self-training loss reduces the impact of classes with many pseudo labeled nodes, which is especially useful to rectify the training process when the predictions in the early epochs are incorrect or less confident, compared with ground truth. 
Further, $\losssst$ only considers high-confident nodes in $\unlabelset'$  defined in Eq. \eqref{eq:betaConfidence}, and filter out low-confidence nodes in the training epochs.
\begin{equation}\label{eq:lossOfOurs}
\begin{aligned}
    \losssst =&\sum_{\forall v_i\in\unlabelset^{'} 
    }\frac{1}{N_i+1}\cdot  \loss(\tilde{\by}_i,\bF_i)
    \end{aligned}
\end{equation} 

Compared with existing techniques \citep{li2018deeper}, our stabilized pseudo labeling technique has major differences. First, we develop the stabilizer to re-weight the importance of pseudo labels in the loss function, so as to address the unstable issue of existing GNNs.
Second, we select only those nodes with high-confidence pseudo labels satisfying $\beta$ threshold, and adaptively update the pseudo labels \textit{per epoch}, meaning that a pseudo label in previous epoch will be removed in the next epoch if its confidence becomes low. On the other hand, existing methods keep a pseudo label once it is selected and never remove it in later epochs, which may harm the training quality if the pseudo label is wrong compared with ground truth.

\subsubsection{Negative Sampling Regularization}\label{sec:negativeSampling}
Under extreme cases when labeled nodes are very few (\eg 1 labeled node per class), we further design a negative sampling regularization technique to improve performance. 
Existing studies mainly use negative sampling to get embeddings in an unsupervised manner \citep{velivckovic2018deep,yang2020understanding,MIAO2022667}. On the contrary, we customize negative sampling for the semi-supervised node classification task, and apply it over \textit{labels} instead of embeddings. 
Intuitively, the label of a node $v$ should be distant to the label of another node $u$ if these two nodes are faraway on the input graph $\graph$.
Specifically, a positive sample is a node $v_i$ in $\labelset$ or $\unlabelset'$. 
We sample a set $\mathcal{I}$ of positive samples from $\labelset\cup\unlabelset'$ uniformly at random.
The negative samples of a positive sample $v_i$ are the nodes that are not directly connected to $v_i$ in graph $\graph$.
For \textit{each} positive sample $v_i$ in $\mathcal{I}$, we sample a fixed-size set $\mathcal{J}_i$ of negative samples uniformly at random. 
For a positive-negative pair $(v_i,v_j)$, compared with the $\tilde{\bY}_i$
of $v_i\in\labelset\cup\unlabelset'$, the intention is to let the output vector $\bF_j$ of $v_j$ to be as different as possible. 
Without ambiguity, here we use the symbol $\tilde{\bY}_i$ to represent the pseudo label for node $v_i$ in $\unlabelset'$ or ground-truth label of node $v_i$ in $\labelset$.
Denote $\mathbf{1}$ as the all-one vector in $\real^c$.
Then the total loss of all positive-negative pairs is
\begin{equation}\label{eq:negloss}
        \loss_{neg}=\sum_{\forall v_i\in\mathcal{I} }\sum_{\forall v_j\in\mathcal{J}_i}\frac{1}{|\mathcal{I}|\cdot|\mathcal{J}_i|}L(\tilde{\by}_i,\mathbf{1} - \bF_{j}).
\end{equation}

\subsubsection{Overall Objective Function}\label{sec:overallalgo}

The overall loss $\loss_{total}$   combines the stabilized pseudo labeling loss and the negative sampling loss in Eq. (\ref{eq:lossOfOurs}) and Eq. (\ref{eq:negloss}) by
\begin{equation}\label{eq:gradientOfours}
\begin{aligned}
    \loss_{total} =&\frac{1}{|\labelset|}\cdot\sum_{\forall v_i\in \labelset} \loss(\by_i,\bF_i)+\lambda_1 \losssst+\lambda_2 \loss_{neg},   
    \end{aligned}
\end{equation}
where $\lambda_1$ and $\lambda_2$ are factors controlling the impact of these two losses.

\begin{algorithm}[!t]
\SetAlgoNoLine
\caption{\ours Framework Over GNNs}\label{algo:main}
\textbf{Input}: Graph $\graph=(\vertice,\edge,\feature)$ with labeled node set $\labelset$ and unlabeled node set $\unlabelset$
\textbf{Output}: the learned classifier $f(\cdot,\theta)$.
Generate initial parameter $\theta$ for model $\classifier(\cdot,\cdot)$.\\
\For{\text{\rm each epoch} $t=0,1,2,...,T$}{
Use GNN to compute {prediction} $\bF\leftarrow f(\graph,\theta)$\\
Get high confidence set $\unlabelset'$ and its stabilizing factor $\frac{1}{N_i+1}$ per node $v_i$(Section \ref{sec:stableselftraining})\\
Get positive samples and corresponding negative samples using $\labelset\cup\unlabelset'$ and $\graph$ (Section \ref{sec:negativeSampling})\\
Get $\loss_{total}$ of current epoch by Eq. (\ref{eq:gradientOfours}) (Section \ref{sec:overallalgo})\\ 
Update {model parameters} by  $\theta\leftarrow \text{Adam Optimizer}(\theta,gradient=\nabla_{\theta} \loss_{total})$. \\

\If {\text{\rm Convergence}}{Break}
}
\end{algorithm}

Algorithm \ref{algo:main} shows the pseudo-code of the \ours framework over GNNs, and it takes as input a graph $\graph$ with labeled nodes $\labelset$ and unlabeled nodes $\unlabelset$. Note that \ours can be instantiated over either a shallow  or a deep GNN, \eg GCN and DAGNN introduced in Section~\ref{sec:relatedwork}.
The output of Algorithm~\ref{algo:main} is the learned classification model $\classifier$ with trainable parameters $\theta$.
At Line 3, \ours initializes the trainable parameters  $\theta$ by Xavier~\citep{glorot2010understanding}. 
Then from Lines 4 to 13, \ours trains the classification model per epoch $t$ iteratively, until convergence or the max number $T$ of epochs is reached. 
We adopt a widely-used early-stopping technique for convergence~\citep{liu2020towards,earlystopping3}. 
Specifically, after half of the max number of epochs (half of $T=1000$ epochs in experiments), given a tolerance duration (100 epochs), if the validation loss of the current epoch is higher than the smallest validation loss of the past 100 epochs in the tolerance duration, the model converges and terminates. The result of the model with the smallest validation loss in the tolerance duration is returned as the final result.
At Line 5, \ours first uses a GNN to obtain the forward prediction output $\bF$. Then at Line 6, \ours detects the pseudo-labeled  set $\unlabelset'$ and obtains the stabilizer $\frac{1}{N_i+1}$ of each node $v_i$ in $\unlabelset'$, after which, at Line 7 we perform negative sampling to obtain positive samples $\mathcal{I}$ and negative samples $\mathcal{J}_i$.
At Line 8, \ours computes loss $\loss_{total}$ of current epoch according to Eq. (\ref{eq:gradientOfours}).
And at Line 9, \ours updates model parameters $\theta$ for next epoch by Adam optimizer~\citep{adam}.

\section{Experiments} \label{sec:experiments}

We evaluate \ours  against 10 competitors for semi-supervised node classification on 5 benchmark graph datasets. All experiments are conducted on a machine powered by  an Intel(R) Xeon(R) E5-2603 v4 @ 1.70GHz CPU, 131GB RAM, 16.04.1-Ubuntu, and 4 Nvidia Geforce 1080ti Cards with Cuda version 10.2.  Source codes of all competitors are obtained from the respective authors.
Our \ours framework is implemented in Python, using libraries including PyTorch~\citep{paszke2019pytorch} and PyTorch Geometric~\citep{fey2019fast}.

\subsection{Implementation Details}
We instantiate our \ours framework over a 2-layer GCN and a  deep DAGNN to demonstrate the effectiveness and applicability of \ours.
The instantiation of \ours over GCN and DAGNN are dubbed as \oursG and \oursD respectively. 
 \oursG and \oursD have parameters (i) inherited from GCN and DAGNN and (ii) developed in \ours.
Hence, we first tune the best parameters of the base models under each classification task setting on each dataset and report this result for them for a fair comparison.

\begin{table}[!t]
    \caption{Dataset Statistics}\label{table:dataset}
    \centering
\small
    \begin{tabular}{l|l|l|l|l|l}
    \toprule
    Dataset &Cora& Citeseer &PubMed &Cora-full & Ogbn-arxiv\\
    \midrule
    \# of Nodes&2708 &3327&19717&19793 &169343 \\
		\# of Edges&5429 &4732&44338&65311  &1166243\\
		\# of Features&1433 &3703&500&8710  &  128 \\
	    \# of Classes&7&6&3&67&  40    \\
    \bottomrule
    \end{tabular}
    \end{table}

\header
{\bf Base models (GCN and DAGNN).} In  the base models, we tune parameters, including $L_2$ regularization rate  with  search space in $\{1e\text{-}2, 5e\text{-}3,1e\text{-}3,$ $5e\text{-}4,1e\text{-}4, 5e\text{-}5,0\}$ and  dropout rate in \{0.5, 0.8\}. For DAGNN, the level $k$ of propagation after MLP is searched in \{10, 15, 20\}. 
The number of hidden units of GCN and MLP (in DAGNN) is 64 units without bias.
The number of layers of GCN and MLP (in DAGNN) is 2 layers. The learning rate of Adam Optimizer is 0.01. The activation function is RELU. The maximum number of training epochs is 1000.
Moreover, early stopping is triggered when the validation loss is smaller than the average validation loss of previous 100 epochs, and the current epoch is beyond 500 epochs.

\header
{\bf \oursG and \oursD.} After finding  the best hyper parameters of the base models, we then tune the parameters in \ours. $\lambda_1$ is searched in $\{0.1, 1\}$ and $\lambda_2$  is searched in $\{0, 0.1,  1\}$.
Stabilizing enabler searches in \{True, False\}. 
The number of positive and negative samples ($|\mathcal{I}|, |\mathcal{J}_i|$) is searched in \{$(1, 10)$, $(2, 5)$, $(5, 2)$, $(10, 1)$\}. For instance, $(2,5)$ means that we sample 2 positive nodes and then for each positive node, we sample 5 negative nodes.

\subsection{Datasets and Competitors}

\header
{\bf Datasets.}
Table \ref{table:dataset} shows the statistics of the 5 real-world  graphs used in our experiments. We list the number of nodes, edges, features and classes in every dataset. 
Specifically, the 5 datasets are Cora~\citep{sen2008collective}, Citeseer~\citep{sen2008collective}, Pubmed~\citep{sen2008collective}, Core-full~\citep{bojchevski2018deep}, and Obgn-arxiv~\citep{ogb}, all of which are widely used  in existing studies \citep{liu2020towards,li2018deeper}.

\header
{\bf Competitors.}
We compare with 10 existing solutions,
including LP (Label Propagation)~\citep{wu2012learning}, 
G2G~\citep{bojchevski2018deep},  DGI~\citep{velivckovic2018deep},  GCN~\citep{kipf2016semi}, GAT~\citep{velickovic2017graph}, 
APPNP~\citep{klicpera2018predict}, DAGNN~\citep{liu2020towards}, STs~\citep{li2018deeper}, LCGCN and LCGAT in \citep{xu2020lcgcn}.
In particular, GCN, GAT, APPNP, and DAGNN are GNNs. 
DGI and G2G are unsupervised network embedding methods. 
STs represents the four variants in~\citep{li2018deeper}, including Self-Training, Co-Training, Union, and Intersection; we summarize the best results among them as the results of STs.
We use the parameters suggested in the original papers of the competitors to tune their models, and report the best results of the competitors. Notice that for unsupervised network embedding methods, including DGI and G2G, after obtaining the  embeddings, we use logistic regression to train a node classifier over the embedding \citep{velivckovic2018deep}.

\begin{table}[t]
\caption{Absolute accuracy improvements (in percentage) of \ours (\ie \oursG and \oursD) over base models GCN and DAGNN on 4 datasets, averaged over 100 random data splits.}\label{tbl:expOverBaseModels}
    \renewcommand\tabcolsep{1pt}
    \centering
    \small
    \begin{tabular}{l|ccccc|ccccc}
    \toprule
    \multicolumn{1}{l|}{\# Labels }      & \multicolumn{5}{c|}{{Cora}}     &    \multicolumn{5}{c}{{CiteSeer}}    \\
    
             per class                                                        & 1                   & 3                   & 5                   & 10                  & 20         & 1                   & 3                   & 5                   & 10                  & 20      \\ \midrule 
             
    {GCN}                                                               & 44.6	&63.8&	71.3&	77.2&	81.4 &40.4&	53.5&	61.0&	65.8&	69.5  \\
    {\oursG}                                                       & {62.5}               & {72.8}              & {75.8}          & {80.7}              & {82.5}     & {56.2}                & {66.4}                & {68.0}          & {70.2}               & {72.1}                       \\
 \textbf{Improvement} &  \textbf{+17.9} &  \textbf{+9.0}&  \textbf{+4.5}&  \textbf{+3.5}&  \textbf{+1.1}&  \textbf{+14.2}&  \textbf{+12.9}&  \textbf{+7.0}&  \textbf{+4.4}&  \textbf{+2.6}
\\
    \midrule
    {DAGNN}                                                             & {59.8}&	{72.4}&	{76.7}&	{80.8}&	{83.7} & {46.5}	&58.8&	63.6&	{67.9}&	71.2        \\ 
    {\oursD}                                                          & {66.4}                & {77.6}               & {79.8}                & {82.2}                & {84.1}     & {48.5}              & {65.9}              & {67.9}    & {69.8}            & {72.1}  \\     
 \textbf{Improvement} &  \textbf{+6.6}&  \textbf{+5.2}&  \textbf{+3.1}&  \textbf{+1.4}&  \textbf{+0.4}&  \textbf{+2.0}&  \textbf{+7.1}&  \textbf{+4.3}&  \textbf{+1.9}&  \textbf{+0.9}\\
    \bottomrule
    \end{tabular}

    \renewcommand\tabcolsep{1.7pt}
    \centering
    \small
    \begin{tabular}{l|ccccc|ccccc}
    \toprule
    \multicolumn{1}{l|}{\# Labels }      &  \multicolumn{5}{c|}{{PubMed}}     &    \multicolumn{5}{c}{{Cora-Full}}     \\
    
             per class                                                        & 1                   & 3                   & 5                   & 10                  & 20         & 1                   & 3                   & 5                   & 10                  & 20      \\ \midrule 
             
    {GCN}                                                               &55.5	&66.0	&70.4	&74.6	&78.7	&24.5	&41.4	&48.1	&{55.8}	&60.2   \\
    {\oursG}                                                       & {60.8}                & {67.8}                & {71.6}          & {76.1}              & {79.4}   & {30.8}              & {44.9}                &   {49.4}&	{56.6}&	{60.9}                        \\
 \textbf{Improvement} &   \textbf{+5.3}&  \textbf{+1.8}&  \textbf{+1.2}&  \textbf{+1.5}&  \textbf{+0.7}&  \textbf{+6.3}&  \textbf{+3.5}&  \textbf{+1.3}&  \textbf{+0.8}&  \textbf{+0.7}
\\
    \midrule
    {DAGNN}                                                           &{59.4}	&{69.5}	&{72.0}	&{76.8}	&{80.1} &27.3	&43.2	&{49.8}	&{55.8}&	60.4           \\ 
    {\oursD}                                                          & {61.0}                & {72.1}             &  {74.9}        & {78.2}                &  {80.6}      & {27.6}                & {44.4}                &  {51.1}&	{56.8}&	{61.2}  \\     
 \textbf{Improvement} &  \textbf{+1.6}&  \textbf{+2.6}&  \textbf{+2.9}&  \textbf{+1.4}&  \textbf{+0.5}&  \textbf{+0.3}&  \textbf{+1.2}&  \textbf{+1.3}&  \textbf{+1.0}&  \textbf{+0.8}\\
    \bottomrule
    \end{tabular}
    \end{table}

\begin{table}[t]
\caption{Accuracy results (in percentage)  on PubMed and Cora-full respectively, averaged over 100 random data splits. ({The \textbf{best} accuracy is in \textbf{bold}.)} }\label{tbl:exp Pubmed Corafull}
\small
\renewcommand\tabcolsep{4pt}
\centering
\begin{tabular}{l|ccccc|ccccc}
\toprule
\multicolumn{1}{l|}{\# Labels    }                                            & \multicolumn{5}{c|}{{PubMed}}                                                                                            & \multicolumn{5}{c}{{Cora-full}}                                                                                                \\
per class                                                             & 1                   & 3                   & 5                   & 10                  & 20         & 1                   & 3                   & 5                   & 10                  & 20     \\ \midrule 

\textbf{\oursG}                                                      & {60.8}                & {67.8}                & {71.6}          & {76.1}              & {79.4}   & \textbf{30.8}              & \textbf{44.9}                &   {49.4}&	{56.6}&	{60.9}                                  \\
\textbf{\oursD}                                                      & \textbf{61.0}                & \textbf{72.1}             &  \textbf{74.9}        & \textbf{78.2}                &  \textbf{80.6}      & {27.6}                & {44.4}                &  \textbf{51.1}&	\textbf{56.8}&	\textbf{61.2} 
\\  
\midrule 
{LP}                                                                & 55.7          & 61.9                & 63.5                & 65.2                & 66.4  & 26.3                & 32.4                & 35.1                & 38.0                & 41.0     \\
{G2G}                                          &55.2	&64.5	&67.4	&72.0	&74.3	&25.8	&36.4	&43.3	&49.3	&54.3     \\
{DGI}                                           & 55.1	&63.4	&65.3	&71.8	&73.9	&26.2	&37.9	&46.5	&55.3	&59.8       \\
{ST}s                                                     & 55.1                & 65.4                & 69.7               & 74.0                & 78.5                &  {29.2}         &  {43.6}         &  48.9          &  53.4          &  {60.8}    \\
{GCN}                                             &55.5	&66.0	&70.4	&74.6	&78.7	&24.5	&41.4	&48.1	&{55.8}	&60.2   \\  

{DAGNN}                              &{59.4}	&{69.5}	&{72.0}	&{76.8}	&{80.1} &27.3	&43.2	&{49.8}	&{55.8}&	60.4          \\
{APPNP}                                               &54.8&	66.9&	70.8&	76.0&	{79.4}   
&   	24.3&	41.5&	48.5&	55.3&	60.1\\  
{GAT}   &   52.7&	64.4&	69.4&	73.7&	73.5   &24.8&	41.0&	47.5&	54.7&	59.9        \\  
{LCGCN}                                             & 56.6	& 69.2	&  72.6	& 74.6	& 80.0	& 26.7&	43.9&	49.2&	55.9&	60.5 \\  
{LCGAT}    &49.5   &  59.2	&62.3& 70.2 &65.3	  &   27.4&	43.2&	48.4&	55.0&	60.1                               \\  
\bottomrule
\end{tabular}
\end{table}

\begin{table}[t]\caption{Accuracy results (in percentage)  on Cora and CiteSeer respectively, averaged over 100 random data splits. ({The \textbf{best} accuracy is in \textbf{bold}.)}}\label{tbl:exp Cora Citeseer}
    \small
    \renewcommand\tabcolsep{4pt}\centering
    \begin{tabular}{l|ccccc|ccccc}
    \toprule
    \multicolumn{1}{l|}{\# Labels }      & \multicolumn{5}{c|}{{Cora}}     &    \multicolumn{5}{c}{{CiteSeer}}      \\
    
             per class                                                        & 1                   & 3                   & 5                   & 10                  & 20         & 1                   & 3                   & 5                   & 10                  & 20     \\ \midrule 
             
    \textbf{\oursG}                                                       & {62.5}               & {72.8}              & {75.8}          & {80.7}              & 82.5     & \textbf{56.2}                & \textbf{66.4}                & {68.0}          & {70.2}               & \textbf{72.1}   \\
    \textbf{\oursD}                                                          & \textbf{66.4}                & \textbf{77.6}               & \textbf{79.8}                & \textbf{82.2}                & \textbf{84.1}     & {48.5}              & {65.9}              & 67.9    & {69.8}            & \textbf{72.1}   \\ 
    \midrule
    {LP}                                                                & 51.5                & 60.5                & 62.5                & 64.2                & 67.3   & 30.1                & 37.0                & 39.3                & 41.9                & 44.8                   \\
    {G2G}                                                                               & 54.5          &  68.1          & 70.9                & 73.8                & 75.8       & 45.1          & 56.4          & 60.3                & 63.1                & 65.7           \\
    {DGI}                                                                             &  55.3          &  70.9          &  72.6          &  76.4          & 77.9             &  46.1          & {59.2}          & {64.1}          &  67.6 & 68.7       \\
    {ST}s                                                                & 53.1                & 67.3               & 72.5                & 76.2                & 79.8         & 37.2                & 51.8                & 60.7                & 67.4                & 70.2          \\
    {GCN}                                                               & 44.6	&63.8&	71.3&	77.2&	81.4 &40.4&	53.5&	61.0&	65.8&	69.5      \\
    {DAGNN}                                                             & {59.8}&	{72.4}&	{76.7}&	{80.8}&	{83.7} & {46.5}	&58.8&	63.6&	{67.9}&	71.2                 \\ 

    {APPNP}                                                             & 44.7&	66.3	&74.1	&79.0 &	81.9  &34.6&	52.2&	59.4&	66.0&	{71.8}           \\ 
    {GAT}                                                               & 41.8                & 61.7                & 71.1                & 76.0                & 79.6      & 32.8                & 48.6                & 54.9                & 60.8                & 68.2                        \\
    
    {LCGCN}                 
    &63.6&	74.4&	77.5&	80.4&	82.4&	55.3&	59.0&	{68.4}	&{70.3}&	{72.1} \\
    {LCGAT}                 
    &58.7&	{74.5}&	{77.5}&	79.7&	{82.6}&{50.9}&	{66.3}&	\textbf{68.5}&	\textbf{70.9}&	71.5\\
    \bottomrule
    \end{tabular}
    \end{table}

\subsection{Experimental Settings}\label{sec:expsetting}
We evaluate our framework and the competitors on semi-supervised node classification tasks with various settings.
In particular, for each graph dataset, we repeat experiments on 100 random data splits as suggested in \citep{liu2020towards,li2018deeper} and report the average performance.
For each graph dataset, we vary the number of labeled nodes per class in $\{1, 3, 5, 10, 20\}$, where $1,3,5$ represent the very few-labeled settings.
Following convention in existing work \citep{liu2020towards}, we explain what a random data split is, as follows.
For example, when the number of labeled nodes per class on Cora is 3 (denoted as Cora-3), since Cora has 7 classes, we randomly pick 3 nodes per class, combining together as a training set of size 21 (\ie the labeled node set $\labelset$), and then, among the remaining nodes, we randomly select 500 nodes as a validation set, and 1000 nodes as a test set. 
Every data split consists of a training set, a validation set, and a test set. 
Classification accuracy is defined as the fraction of the testing nodes with class labels correctly predicted by the learned classifier.

\subsection{Overall Experimental Results}

In Table \ref{tbl:expOverBaseModels}, we first report the absolute improvements of our \ours  applied over base models GCN and DAGNN, when varying the number of labeled nodes per class in $\{1,3,5,10,20\}$. The accuracy performance of \oursG (resp. GCN) and \oursD (resp. DAGNN) on each of the 4 datasets are obtained by averaging over 100 random data splits.
The overall observation is that \oursG and \oursD consistently achieve superior performance compared with their respective base models GCN and DAGNN, often by a significant margin, which validates the power of the proposed \ours framework, to be readily applicable to boost the performance of existing GNNs.
For instance, on Cora-1, \oursG has accuracy $62.5\%$, while the accuracy of GCN is $44.6\%$, which indicates that \ours improves GCN by $17.9\%$. Also, on Cora-1, \oursD improves DAGNN by $6.6\%$.
Moreover, observe that when labels are sufficient (\eg Cora-20 and CiteSeer-20 in Table \ref{tbl:expOverBaseModels}), our methods \oursG and \oursD are still better than GCN and DAGNN respectively, further validating the effectiveness of the proposed \ours, by stabilizing the training process as shown in Figures \ref{fig:dist}(c) and \ref{fig:dist}(d) of Section \ref{sec:intro}.

Table \ref{tbl:exp Pubmed Corafull} reports the  classification accuracy (in percentage) of all methods on  PubMed and Cora-full when varying the number of labeled nodes per class in $\{1, 3, 5, 10, 20\}$. 
The first two rows are the performance of \oursG and \oursD, while the remaining rows are the performance of the 10 competitors.
On PubMed, \oursD consistently achieves the highest accuracy among all methods under all settings in $\{1,3,5,10,20\}$.
For instance, on PubMed-3, \oursD has accuracy $72.1\%$, while the best competitor's performance is $69.5\%$.
Further, on Cora-full in Table \ref{tbl:exp Pubmed Corafull}, \oursG and \oursD achieve the highest accuracy under the settings $\{1,3\}$ and $\{5, 10,20\}$ respectively (in bold), compared with all competitors. 
The results in Table \ref{tbl:exp Pubmed Corafull} demonstrate the effectiveness of our \ours framework to boost classification performance over existing methods. 
Table \ref{tbl:exp Cora Citeseer} reports the classification results of all methods on Cora and CiteSeer under all settings in $\{1, 3, 5, 10, 20\}$.
On Cora, observe that \oursD obtains the highest accuracy under all settings. For instance, on Cora-1, \oursD has accuracy $66.4\%$, while that of the best competitor LCGCN is $63.6\%$.
On CiteSeer in Table \ref{tbl:exp Cora Citeseer}, \oursG has the best performance on CiteSeer-1 and CiteSeer-3 and CiteSeer-20, while achieving similar performance compared with LCGCN and LCGAT on CiteSeer-5 and CiteSeer-10.
Summing up, the proposed SST framework is effective to boost classification accuracy as validated in  Tables  \ref{tbl:exp Pubmed Corafull} and \ref{tbl:exp Cora Citeseer},  especially under the situation when very few labeled nodes are available.

\begin{table*}[!t]
\caption{Accuracy results (in percentage)  on {Ogbn-arxiv}, averaged over 100 random data splits. (The \textbf{best} accuracy is in \textbf{bold}.) }\label{tbl:exp ogbn-arxiv}

\small
\centering

\begin{tabular}{l|ccccc}
\toprule
\multicolumn{1}{l|}{\# Labels    }                                            & \multicolumn{5}{c}{{Ogbn-arxiv}}                                                                                                               \\
per class                                                             & 1                   & 3                   & 5                   & 10                  & 20     \\ \midrule 

\textbf{\oursG}   &  25.57	&41.12&	44.42 	&50.66&	53.99       \\
\textbf{\oursD}  & \textbf{27.20}&	\textbf{42.66} &	\textbf{47.29}&	\textbf{53.76}&	\textbf{57.13}       
\\  
\midrule 
{DGI}& 15.25 &	24.35 &	31.89 &	38.55&	43.34  \\
{ST}s&  22.49	& 34.44	&40.57	&49.14	&52.23     \\      
{GCN} &  25.11 	&36.32 	&40.90&	48.42 	&53.76  \\

{DAGNN} & 25.18 &	40.72	&46.66 &	52.57 	&56.07   \\
{APPNP} & 25.21&	39.74 	&45.58 &	49.51&53.50  \\
{GAT} & 25.86&	38.52	&45.58&	50.83 	&53.53  \\
{LCGCN}& 25.96&	38.62	&44.68 &	52.06 	&57.09 \\
{LCGAT} &    26.93&	37.31 	&43.09 	&47.72 &	55.05            \\  
\bottomrule
\end{tabular}
\end{table*}

\begin{table*}[!t]
\caption{Total training time in seconds (s) and training time per epoch in milliseconds (ms) on {Ogbn-arxiv}. }\label{tbl:exp ogbn-arxiv time}

\centering

\small

\setlength{\tabcolsep}{4pt}
\begin{tabular}{l|ccccc|ccccc}
\toprule
\multicolumn{1}{l|}{\# Labels    }                                            & \multicolumn{5}{c|}{{ Total training time (s) }}                                         & \multicolumn{5}{c}{{ Training time per epoch (ms)}}                                                                                           \\
per class                                                             & 1                   & 3                   & 5                   & 10                  & 20                                                           & 1                   & 3                   & 5                   & 10                  & 20   \\ \midrule    

\textbf{\oursG}   & 87.4	&90.8&105.7	&92.8&	107.3&  158.7 &171.6&219.7&180.8&205.7  \\
\textbf{\oursD}  &144.1&218.7&134.7&119.3&	159.1  &257.1 &256.7 & 234.6&156.6 & 187.4    
\\  
\midrule 
{DGI}& 95.0 & 95.0 &	96.8  &	94.1 & 94.3 & 190.7 &	  189.9 &  190.4	 &   190.6&   190.4 \\
{STs}&   46.1 	&  45.4 	& 51.3 	& 52.6 	& 30.0   &55.0 &55.1 & 55.3& 55.5&55.4   \\      
{GCN} & 32.4	&54.5	&56.3&56.3	&56.5   &53.0 & 53.4&53.0 &53.1 &53.4 \\

{DAGNN} & 83.3 &	 110.3 	& 122.2 &	 158.2 	& 156.6     &150.9 & 150.8&150.7 &151.8 & 148.3 \\
{APPNP} &  61.1 &	 71.3 	& 107.8 &	 120.0 	& 119.6   &113.6&113.1 &113.3&113.5& 113.2\\
{GAT} & 93.8 & 94.1 	& 94.0 & 93.7 	& 94.5   &89.9 &90.3 &90.1& 89.7 & 90.5 \\
{LCGCN}&  83.5 &	 84.1 	& 84.6 & 84.9 	& 85.0     &80.61 &80.5&80.6 &80.7 & 80.9 \\
{LCGAT} &  96.7 & 97.2 	& 97.3 	&97.3 &	 97.4      & 93.1&93.5 &93.6 &93.5 &  93.1     \\  
\bottomrule
\end{tabular}
\end{table*}

Table~\ref{tbl:exp ogbn-arxiv}
reports the  classification accuracy (in percentage) of all methods on  Ogbn-arxiv,  when varying the number of labeled nodes per class in $\{1, 3, 5, 10, 20\}$. 
The first two rows are the performance of \oursG and \oursD, while the remaining rows are the performance of the competitors.
\oursD consistently achieves the highest accuracy among all methods under all settings.
For instance, on Obgn-arxiv-3, \oursD has accuracy $42.66\%$, while the best competitor's performance is $40.72\%$.
Further, comparing with the corresponding base models, \oursG (resp. \oursD) is consistently better than GCN (resp. DAGNN).
The results in Table~\ref{tbl:exp ogbn-arxiv}  demonstrate the effectiveness of our \ours framework to improve classification performance.

Note that our \ours framework is integrated into existing GNN models (GCN and DAGNN). Therefore, \oursG and \oursD inherit the efficiency of their corresponding base models, with moderate extra overheads to perform the stabilized self-training techniques proposed in this paper. 
We report the total training time and training time per epoch of all methods on Obgn-arxiv in Table~\ref{tbl:exp ogbn-arxiv time}, to show that the methods are able to scale to large graphs. 
Since different models converge at different epochs, their total training time varies. The total training time of all models are affordable (within 4 minutes). For instance, on Ogbn-arxiv-10, \oursD converges in 119.3 seconds, while DAGNN costs 158.2 seconds and APPNP needs 120.0 seconds. On Ogbn-arxiv-1, \oursG converges in 87.4 seconds, while GAT and LCGAT need 93.8 and 96.7 seconds respectively. 
In terms of training time per epoch, compared with base models (\eg DAGNN), \oursD has similar training time per epoch with moderate extra overheads.
In summary, our methods consistently achieve the highest effectiveness, while having similar efficiency and scalability performance compared with existing methods.

\begin{figure}[!t]
\centering
\begin{tikzpicture}
    \begin{customlegend}[legend columns=4,
        legend entries={\textcolor{black}{ base},\textcolor{black}{base+N}, \textcolor{black}{base+S},\textcolor{black}{base+\ours}},
        legend style={at={(0.45,1)},anchor=north,draw=none,font=\small,column sep=0.15cm}]
    \addlegendimage{line width=0.25mm,color=red,mark=triangle*}
    \addlegendimage{line width=0.25mm,color=violet,mark=pentagon}
    \addlegendimage{line width=0.25mm,color=magenta,mark=o}
    \addlegendimage{line width=0.25mm,color=cyan,mark=x}

    legend columns=2,
        legend entries={},
        legend style={at={(0.45,1.5)},anchor=north,draw=none,font=\footnotesize,color=black,column sep=0.15cm},
        
    \end{customlegend}
\end{tikzpicture}
\\[-\lineskip]
\vspace{-2mm}
\subfloat[{\em Ablation on \oursG with base GCN}]{
\begin{tikzpicture}[scale=1]
    \begin{axis}[
        height=\columnwidth*0.38,
        width=\columnwidth*0.45,
        xlabel=Number of labels per class,
        ylabel={\em Accuracy(\%)},
        xmin=1, xmax=5,
        ymin=42, ymax=85,
        xtick={1,2,3,4,5},
        xticklabel style = {font=\footnotesize,color=black},
        xticklabels={1,3,5,10,20},
        ytick={40,50,60,70,80},
        yticklabels={40,50,60,70,80},
        scaled y ticks = false,
        every axis y label/.style={at={(current axis.north west)},right=3mm,above=0mm},
        label style={font=\footnotesize,color=black},
        tick label style={font=\footnotesize,color=black},
    ]
    \addplot[line width=0.25mm,color=red,mark=triangle*] 
        plot coordinates {
(1,	    44.6)			
(2,		63.8	)
(3,		71.3	)
(4,     77.2	)
(5,     81.4	)
        };    
    
        
    \addplot[line width=0.25mm,color=magenta,mark=o] 
        plot coordinates {
(1,	60.2	)		
(2,	72.0	)
(3,	74.8	)
(4,	79.1	)
(5,	82.5	)
        };
    \addplot[line width=0.25mm,color=violet,mark=pentagon] 
        plot coordinates {
(1,	53.1	)
(2,	67.3	)				
(3,	71.0	)
(4,	77.2	)
(5,	80.6	)
        };
    \addplot[line width=0.25mm,color=cyan,mark=x] 
        plot coordinates {
(1,	62.5	)			
(2,	72.8	)
(3,	75.8	)
(4,	79.5	)
(5,	81.8	)
        };
    \end{axis}
\end{tikzpicture}\label{fig:ablationG}
}%
\hspace{4.5mm}
%
%
\subfloat[{\em Ablation on \oursD with base DAGNN}]{
\begin{tikzpicture}[scale=1]
    \begin{axis}[
        height=\columnwidth*0.38,
        width=\columnwidth*0.45,
        xlabel=Number of labels per class,
        ylabel={\em Accuracy(\%)},
        xmin=1, xmax=5,
        ymin=58, ymax=85,
        xtick={1,2,3,4,5},
        xticklabel style = {font=\footnotesize,color=black},
        xticklabels={1,3,5,10,20},
        ytick={60,70,80},
        yticklabels={60,70,80},
        scaled y ticks = false,
        every axis y label/.style={at={(current axis.north west)},right=3mm,above=0mm},
        label style={font=\footnotesize,color=black},
        tick label style={font=\footnotesize,color=black},
    ]
    \addplot[line width=0.25mm,color=red,mark=triangle*] 
        plot coordinates {
(1,	    59.8)				
(2,		74.1	)
(3,		77.6	)
(4,     80.9	)
(5,     83.7	)
        };    
    
        
    \addplot[line width=0.25mm,color=magenta,mark=o] 
        plot coordinates {
(1,	65.4	)				
(2,	76.7	)
(3,	79.5	)
(4,	81.9	)
(5,	84.1	)
        };
    \addplot[line width=0.25mm,color=violet,mark=pentagon] 
        plot coordinates {
(1,	61.5	)				
(2,	74.9	)
(3,	77.7)
(4,	80.6	)
(5,	83.3	)
        };
    \addplot[line width=0.25mm,color=cyan,mark=x] 
        plot coordinates {
(1,	66.4	)				
(2,	77.6	)
(3,	79.8	)
(4,	81.8	)
(5,	83.9	)
        };
    \end{axis}
\end{tikzpicture}\hspace{0.5mm}\label{fig:ablationD}
}%

\caption{Ablation study of \ours on Cora.} \label{fig:ablation}
\vspace{-2mm}
\end{figure}

\subsection{Ablation Study} \label{sec:ablationandparameter}
We conduct ablation study to evaluate the contributions of the techniques of \ours presented in Section~\ref{sec:algo}.
In particular, let base models be either GCN or DAGNN.
Then denote base+\ours as the method with the whole \ours framework enabled (\ie \oursG or \oursD), base+S as the method with only stabilized pseudo labeling loss in Eq. (\ref{eq:lossOfOurs}) enabled, and base+N as the method with only negative sampling loss in Eq. (\ref{eq:negloss}) enabled.
Figures \ref{fig:ablationG} and \ref{fig:ablationD} report the ablation results of \oursG and \oursD respectively, on Cora when varying the number of labels per class in $\{1,3,5,10,20\}$.
Observe that the accuracy of base+S and base+N are always better than the base models, \ie GCN and DAGNN, indicating the effectiveness of the proposed pseudo labeling and negative sampling techniques in \ours. 
Further, the accuracy of base+\ours is always the highest under almost all settings.
The ablation study demonstrates the power of our proposed techniques to improve classification accuracy.

We further vary $\beta$ from 0.1 to 0.9, and report the performance of \oursG in Figure \ref{fig:ablationBetaSSTGCN}, for node classification on Cora-5, CiteSeer-3, PubMed-1, and Cora-full-3.
The corresponding number of pseudo labels when varying $\beta$ in \oursG is reported in Table \ref{tab:ablationBetaSSTGCN}.
As $\beta$ increases, the number of pseudo labels decreases (Table \ref{tab:ablationBetaSSTGCN}).
For \oursG, the best result is achieved at $\beta=0.6,0.2,0.5,0.3$ for  Cora-5, CiteSeer-3, PubMed-1, and Cora-full-3 respectively as shown in Figure \ref{fig:ablationBetaSSTGCN}, which exhibits the trade off between the confidence and the number of pseudo labels decided by Eq. (\ref{eq:betaConfidence}). 
Figure \ref{fig:ablationBetaSSTDA} and Table \ref{tab:ablationBetaSSTDAGNN} report the accuracy of \oursD and the number of pseudo labels respectively when varying $\beta$. 
The best result is achieved at $\beta=0.6,0.4,0.3,0.3$ for node classification on Cora-5, CiteSeer-3, PubMed-1, 
and Cora-full-3 respectively. 
In Table \ref{tab:ablationBetaSSTDAGNN}, the number of pseudo labels decreases as $\beta$ increases.

\begin{figure}[!t]
\centering
\begin{tikzpicture}
    \begin{customlegend}[legend columns=4,
        legend entries={
        \textcolor{black}{ Cora-5},
        \textcolor{black}{CiteSeer-3}, 
        \textcolor{black}{PubMed-1},
        \textcolor{black}{Cora-full-3}
        },
        legend style={at={(0.45,1)},anchor=north,draw=none,font=\small,column sep=0.15cm}]
    \addlegendimage{line width=0.25mm,color=red,mark=triangle*}
    \addlegendimage{line width=0.25mm,color=magenta,mark=o}
    \addlegendimage{line width=0.25mm,color=violet,mark=pentagon}
    \addlegendimage{line width=0.25mm,color=cyan,mark=x}

    legend columns=2,
        legend entries={},
        legend style={at={(0.45,1.5)},anchor=north,draw=none,font=\footnotesize,color=black,column sep=0.15cm},
        
    \end{customlegend}
\end{tikzpicture}
\\[-\lineskip]
\vspace{-2mm}
 \subfloat[Vary $\beta$ in \oursG]{
\begin{tikzpicture}[scale=1]
    \begin{axis}[
        height=\columnwidth*0.38,
        width=\columnwidth*0.45,
        ylabel={\em Accuracy(\%)},
        xmin=1, xmax=9,
        ymin=40, ymax=80,
        xtick={1,2,3,4,5,6,7,8,9},
        xticklabel style = {font=\footnotesize,color=black},
        xticklabels={0.1,0.2,0.3,0.4,0.5,0.6,0.7,0.8,0.9},
        ytick={40,50,60,70,80},
        yticklabels={40,50,60,70,80},
        scaled y ticks = false,
        every axis y label/.style={at={(current axis.north west)},right=3mm,above=0mm},
        label style={font=\footnotesize,color=black},
        tick label style={font=\footnotesize,color=black},
    ]
    \addplot[line width=0.25mm,color=red,mark=triangle*] 
        plot coordinates {
(1.0,74.41243366186505)
(2.0,74.6095526914329)
(3.0,72.60045489006822)
(4.0,74.97346474601972)
(5.0,73.93479909021987)
(6.0,75.70887035633056)
(7.0,74.579226686884)
(8.0,70.08339651250948)
(9.0,69.55269143290371)
        };    
        
    \addplot[line width=0.25mm,color=magenta,mark=o] 
        plot coordinates {
(1.0,63.90154968094804)
(2.0,64.60650258280158)
(3.0,61.93254329990885)
(4.0,63.50349437860834)
(5.0,61.24886052871468)
(6.0,61.51625645700395)
(7.0,57.43542996049833)
(8.0,49.28289273776968)
(9.0,54.1750227894257)
        };
    \addplot[line width=0.25mm,color=violet,mark=pentagon] 
        plot coordinates {
(1.0,52.915884531479875)
(2.0,58.610927908274554)
(3.0,52.04961696514636)
(4.0,55.45964182436202)
(5.0,61.41291664552788)
(6.0,58.54218456699306)
(7.0,55.00253665465984)
(8.0,55.8528232966364)
(9.0,55.43097762670589)
        };
    \addplot[line width=0.25mm,color=cyan,mark=x] 
        plot coordinates {
(1.0,44.68641924327672)
(2.0,43.754193981314195)
(3.0,45.528312600010324)
(4.0,43.000567800547145)
(5.0,41.94962060599804)
(6.0,41.82470448562432)
(7.0,42.51432405925774)
(8.0,42.439993805812215)
(9.0,41.610488824652876)
        };
    \end{axis}
\end{tikzpicture} \label{fig:ablationBetaSSTGCN}
}
\subfloat[Vary $\beta$ in \oursD]{
\begin{tikzpicture}[scale=1]
    \begin{axis}[
        height=\columnwidth*0.38,
        width=\columnwidth*0.45,
        ylabel={\em Accuracy(\%)},
        xmin=1, xmax=9,
        ymin=40, ymax=80,
        xtick={1,2,3,4,5,6,7,8,9},
        xticklabel style = {font=\footnotesize,color=black},
        xticklabels={0.1,0.2,0.3,0.4,0.5,0.6,0.7,0.8,0.9},
        ytick={40,50,60,70,80},
        yticklabels={40,50,60,70,80},
        scaled y ticks = false,
        every axis y label/.style={at={(current axis.north west)},right=3mm,above=0mm},
        label style={font=\footnotesize,color=black},
        tick label style={font=\footnotesize,color=black},
    ]
    \addplot[line width=0.25mm,color=red,mark=triangle*] 
        plot coordinates {
(1.0,76.59780136467022)
(2.0,76.97498104624715)
(3.0,76.96739954510994)
(4.0,76.69446550416983)
(5.0,76.90674753601213)
(6.0,79.38589840788475)
(7.0,79.42380591357089)
(8.0,77.42228961334344)
(9.0,74.92039423805913)
        };    
        
    \addplot[line width=0.25mm,color=magenta,mark=o] 
        plot coordinates {
(1.0,65.54542692190822)
(2.0,65.05621391674265)
(3.0,65.4846551200243)
(4.0,66.2397447584321)
(5.0,63.1950774840474)
(6.0,63.719234275296266)
(7.0,62.125493770890294)
(8.0,58.195077484047395)
(9.0,57.86539045882711)
        };
    \addplot[line width=0.25mm,color=violet,mark=pentagon] 
        plot coordinates {
(1.0,56.4834356)
(2.0,57.31850240000001)
(3.0,61.5029679)
(4.0,56.299020899999995)
(5.0,57.4062706)
(6.0,49.619755500000004)
(7.0,49.6172188)
(8.0,51.3228654)
(9.0,59.6694739)
        };
    \addplot[line width=0.25mm,color=cyan,mark=x] 
        plot coordinates {
(1.0,45.5683167)
(2.0,45.0766531)
(3.0,46.4187271)
(4.0,45.6756827)
(5.0,46.0354617)
(6.0,46.08166)
(7.0,46.2367728)
(8.0,44.2288236)
(9.0,42.9572085)
        };
    \end{axis}
\end{tikzpicture} \label{fig:ablationBetaSSTDA}
}

 \vspace{-3mm}
\caption{Vary $\beta$. Results are averaged over 20 runs.   } 
\end{figure}

\begin{table}[!t]
    \caption{The number of pseudo labels when varying $\beta$ in \oursG, averaged over 20 runs.}
    \label{tab:ablationBetaSSTGCN}
    \centering
    \vspace{-2mm}
    \small
    \setlength{\tabcolsep}{4pt}
    \begin{tabular}{c|ccccccccc}
    \toprule
\diagbox{Task}{$\beta$}&0.1&0.2&0.3&0.4&0.5&0.6&0.7&0.8&0.9\\
         \midrule
Cora-5&2708&2707&2630&2368&2013&1714&1235&855&369\\
CiteSeer-3&3327&3327&3325&3307&3247&3118&2907&2436&1535\\
PubMed-1&19717&19717&19717&19250&16463&12349&9246&5307&1871\\
Cora-full-3&19789&19704&19366&18879&18006&17041&15553&13616&10233\\
\bottomrule
    \end{tabular}

\end{table}

\begin{table}[!t]
    \caption{The number of pseudo labels when varying $\beta$ in \oursD, averaged over 20 runs.}
    \label{tab:ablationBetaSSTDAGNN}
    \centering
    \vspace{-2mm}
    \small
    \setlength{\tabcolsep}{4pt}
    \begin{tabular}{c|ccccccccc}
    \toprule
\diagbox{Task}{$\beta$}&0.1&0.2&0.3&0.4&0.5&0.6&0.7&0.8&0.9\\
         \midrule
Cora-5&2707&2707&2675&2535&2233&1994&1678&1346&817 \\
CiteSeer-3&3326&3327&3290&3092&2721&2305&1843&1244&575\\
PubMed-1&19717&19717&19717&19691&19472&18674&17733&16219&13843\\
Cora-full-3&19741&
18892&
16976&
14678&
12235&
9914&
7475&
5258&
2914
\\
\bottomrule
    \end{tabular}
\vspace{-2mm}
\end{table}

\section{Conclusion}\label{sec:conclusion}

This paper presents Stabilized Self-Training  (\ours), an effective framework for semi-supervised node classification on few-labeled graph data.
We identify that existing GNNs are with unstable performance under few-labeled settings. We also conduct extensive empirical and theoretical analysis to provide solid explanations for the observations.
\ours is designed with the consideration of the analysis, and achieves superior performance on graphs with extremely few labeled nodes.
\ours consists of a stabilized pseudo labeling technique and a negative sampling regularizer over pseudo labels.
The effectiveness of \ours is  evaluated via extensive experiments.
In the future, we plan to enhance \ours  by investigating other unsupervised techniques, and also implement \ours on top of more GNN architectures to further demonstrate its applicability.

\appendix

\section{Proof} \label{secapp:proof}

\subsection{Proof of Eq. \eqref{eq:boundofgradient}}\label{proof:boundofgradient}


\begin{proof} The first inequality of Eq.~\eqref{eq:boundofgradient} is from the property of conditional expectation. Specifically, for a random variable $X=\nabla_{\theta} \loss(\tilde{\by}_i,\bF_i)- \nabla_{\theta} \loss(\bY_i,\bF_i)$ and a partition of sampling space into two parts: $$A=\{v_i\sim\bU(\unlabelset)|\tilde{\by}_i\neq \by_i\}$$ 
$$B=\{v_i\sim\bU(\unlabelset)|\tilde{\by}_i= \by_i\}$$

Then the following holds since $\EE [X|B]=0$: 
\begin{equation*}
\begin{aligned}
    \EE [X]= &\EE [X|A]\PP(A)+\EE [X|B]\PP(B)\\
    =&\EE [X|A]\PP(A)
\end{aligned}
\end{equation*}

Based on Jensen's inequality, the first inequality of Eq. \eqref{eq:boundofgradient} holds.
\begin{equation*}
    \begin{aligned}
        \EE [X]=&\norm{\EE [X|A]\PP(A)}\\
        \le & \PP(A)\EE[\norm{ X}|A].
    \end{aligned}
\end{equation*}


The second inequality is from the bounded gradient norm assumption and the triangle property of the norm, such that 
\begin{equation*}
\begin{aligned}
    \norm{\nabla_{\theta} \loss(\tilde{\by}_i,\bF_i)- \nabla_{\theta} L(\by_i,\bF_i)}\le & \norm{\nabla_{\theta} \loss(\tilde{\by}_i,\bF_i)}+\norm{ \nabla_{\theta} L(\by_i,\bF_i)}  \\
\le & 2\Theta
\end{aligned}
\end{equation*}

Then we have 
\begin{equation*}
\begin{aligned}
    \frac{|\unlabelset|\cdot \PP(A)}{|\labelset|}\cdot\EE\left[\norm{X}|B\right]  
        \le &  \frac{2\Theta|\unlabelset|}{|\labelset|}\cdot \PP(A) \\
        = & \frac{2\Theta|\unlabelset|}{|\labelset|}\cdot \PP_{v_i\sim \bU(\unlabelset)}(\tilde{\by}_i\neq \by_i)
\end{aligned}
\end{equation*}

Thus Eq.~\eqref{eq:boundofgradient} holds.
\end{proof}

\subsection{Proof of Lemma \ref*{lem:error_of_unbalance}}\label{proof:Lemma1}

The proof of Lemma \ref{lem:error_of_unbalance} needs Lemma \ref{lem:maxbalancegroundtruth} and \ref{lem:maxunbalancepseudo} that are presented and proved as follows. 

\begin{lemma}\label{lem:maxbalancegroundtruth}
    If the ground-truth distribution of labels is $\rho$-balanced, the maximal probability of a class will be bounded as 
    \begin{equation*}
        \begin{aligned}
        \max_{y\in\bC}\PP_{v_i\sim\bU(\unlabelset)}(\by_i=y)\leq \frac{\rho}{\rho+|\bC|-1}.
        \end{aligned}
    \end{equation*}
    \end{lemma}
    \begin{proof}[Proof of Lemma~\ref{lem:maxbalancegroundtruth}]
    One can easily derive from Eq.~\eqref{eq:balancetruth} that for any $y\in\bC$, we have
    \begin{equation*}
        \begin{aligned}
            \PP_{v_i\sim\bU(\unlabelset)}(\by_i=y)\geq \frac{1}{\rho}\PP_{v_i\sim\bU(\unlabelset)}(\by_i=y'').
        \end{aligned}
    \end{equation*}
By the property of probability, denote $y''=\argmax_{y\in\bC}\PP_{v_i\sim\bU(\unlabelset)}(\by_i=y)$ 
    \begin{equation*}
        \begin{aligned}
        1=&\sum_{y\in\bC}\PP_{v_i\sim\bU(\unlabelset)}(\by_i=y)\\
        =&\PP_{v_i\sim\bU(\unlabelset)}(\by_i=y'')+\sum_{y\in\bC,y\neq y''}\PP_{v_i\sim\bU(\unlabelset)}(\by_i=y)\\
        \geq&\PP_{v_i\sim\bU(\unlabelset)}(\by_i=y'')+\frac{1}{\rho}\sum_{y\in\bC,y\neq y''}\PP_{v_i\sim\bU(\unlabelset)}(\by_i=y'')\\
        =& (1+\frac{|\bC|-1}{\rho})\cdot\PP_{v_i\sim\bU(\unlabelset)}(\by_i=y'').
        \end{aligned}
    \end{equation*}
    Then we have 
    \begin{equation*}
        \begin{aligned}
        \max_{y\in\bC}\PP_{v_i\sim\bU(\unlabelset)}(\by_i=y)\leq \frac{\rho}{\rho+|\bC|-1}.
        \end{aligned}
    \end{equation*}
    \end{proof}

    \begin{lemma}\label{lem:maxunbalancepseudo}
    If the distribution of pseudo labels is $\eta$-imbalanced, we have the lower bound of the maximal probability of a class as
    \begin{equation*}
        \begin{aligned}
        \max_{y\in\bC}\PP_{v_i\sim\bU(\unlabelset)}(\tilde{\by}_i=y)\geq \frac{\eta}{\eta+|\bC|-1}.
        \end{aligned}
    \end{equation*}
    \end{lemma}

    \begin{proof}[Proof of lemma~\ref{lem:maxunbalancepseudo}]
    One can easily get from Eq.\eqref{eq:unbalancepseudo} that for any $y\in\bC$, 
    \begin{equation*}
        \PP_{v_i\sim\bU(\unlabelset)}(\tilde{\by}_i=y)\leq \frac{1}{\eta}\PP_{v_i\sim\bU(\unlabelset)}(\tilde{\by}_i=y').
    \end{equation*}
    Due to the property of probability, 
    \begin{equation*}
        \begin{aligned}
        1=&\sum_{y\in\bC}\PP_{v_i\sim\bU(\unlabelset)}(\tilde{\by}_i=y)\\
        =&\PP_{v_i\sim\bU(\unlabelset)}(\tilde{\by}_i=y')+\sum_{y\in\bC,y\neq y'}\PP_{v_i\sim\bU(\unlabelset)}(\tilde{\by}_i=y)\\
        \leq & \PP_{v_i\sim\bU(\unlabelset)}(\tilde{\by}_i=y')+\frac{|\bC|-1}{\eta}\PP_{v_i\sim\bU(\unlabelset)}(\tilde{\by}_i=y').
        \end{aligned}
    \end{equation*}
    Thus $
        \PP_{v_i\sim\bU(\unlabelset)}(\tilde{\by}_i=y')=\max_{y\in\bC}\PP_{v_i\sim\bU(\unlabelset)}(\tilde{\by}_i=y)\geq \frac{\eta}{\eta+|\bC|-1}.$    \end{proof}

    After getting Lemma \ref{lem:maxbalancegroundtruth} and \ref{lem:maxunbalancepseudo}, in the following, we prove Lemma \ref{lem:error_of_unbalance}.

\begin{proof}[Proof of lemma~\ref{lem:error_of_unbalance}]
  We expand the classification error  by the pseudo labels
    \begin{equation}\label{eq:mid}
        \begin{aligned}
            \PP_{v_i\sim\bU(\unlabelset)}(\tilde{\by}_i\neq \by_i)
            =&\sum_{y\in\bC} [\PP_{v_i\sim\bU(\unlabelset)}(\tilde{\by}_i\neq\by_i,\tilde{\by}_i=y)].
        \end{aligned}
    \end{equation}
    Let $y'=\argmax_{y\in\bC}\PP_{v_i\sim\bU(\unlabelset)}(\tilde{\by}_i=y)$, we have 

    \begin{equation*}
        \begin{aligned}
            \sum_{y\in\bC} [\PP_{v_i\sim\bU(\unlabelset)}(\tilde{\by}_i\neq\by_i,\tilde{\by}_i=y)]&\geq \PP_{v_i\sim\bU(\unlabelset)}(\tilde{\by}_i\neq\by_i,\tilde{\by}_i=y')\\
            &= \sum_{y\in\bC,y\neq y'}\PP_{v_i\sim\bU(\unlabelset)}(\by_i=y,\tilde{\by}_i=y')\\
            &\geq \sum_{y\in\bC,y\neq y'}[\PP_{v_i\sim\bU(\unlabelset)}(\by_i=y)\cdot\PP_{v_i\sim\bU(\unlabelset)}(\tilde{\by}_i=y')]\\
            &= [1-\PP_{v_i\sim\bU(\unlabelset)}(\by_i=y')]\cdot\PP_{v_i\sim\bU(\unlabelset)}(\tilde{\by}_i=y').
        \end{aligned}
    \end{equation*}

    We are interested in the max and min values of $\PP_{v_i\sim\bU(\unlabelset)}(\tilde{\by}_i=y')$ and $\PP_{v_i\sim\bU(\unlabelset)}(\by_i=y')$ respectively. 
    From Lemma~\ref{lem:maxbalancegroundtruth}, we have 
    \begin{equation*}
        \begin{aligned}
        \PP_{v_i\sim\bU(\unlabelset)}(\by_i=y')\leq& \max_{y\in\bC}\PP_{v_i\sim\bU(\unlabelset)}(\by_i=y)\\
        \leq&\frac{\rho}{\rho+|\bC|-1}.
        \end{aligned}
    \end{equation*}
    From Lemma~\ref{lem:maxunbalancepseudo}, we can directly derive 
    \begin{equation*}
        \PP_{v_i\sim\bU(\unlabelset)}(\tilde{\by}_i=y')\geq \frac{\eta}{\eta+|\bC|-1}.
    \end{equation*}
    Thus we can get
    \begin{equation*}
        \begin{aligned}
        \PP_{v_i\sim\bU(\unlabelset)}(\tilde{\by}_i=y')-\PP_{v_i\sim\bU(\unlabelset)}(\by_i=y')\geq& \frac{\eta}{\eta+|\bC|-1}(1-\frac{\rho}{\rho+|\bC|-1})\\
        \geq &\frac{\eta}{\eta+|\bC|-1}-\frac{\rho}{\rho+|\bC|-1}.
        \end{aligned}
    \end{equation*}
    Combining all above, we have the lower bound of  classification error
    \begin{equation*}
        \PP_{v_i\sim\bU(\unlabelset)}(\tilde{\by}_i\neq \by_i)\geq \frac{\eta}{\eta+|\bC|-1}-\frac{\rho}{\rho+|\bC|-1}.
    \end{equation*}
    \end{proof}

\section*{CRediT authorship contribution statement}
\noindent\textbf{Ziang Zhou}: Conceptualization, Investigation, Validation, Formal analysis, Writing - Original
Draft. \textbf{Jieming Shi}: Resources, Supervision, Methodology, Funding acquisition, Writing - Review \& Editing.
\textbf{Shengzhong Zhang}: Software, Data Curation. \textbf{Zengfeng Huang}: Conceptualization, Supervision, Resources. \textbf{Qing
Li}: Supervision, Funding acquisition.

\section*{Acknowledgments}
This work is supported by  Hong Kong RGC ECS (No. 25201221), RGC GRF (No. 15200021), and PolyU Start-up Fund (P0033898).
Zengfeng Huang is supported by National Natural Science Foundation of China No. 62276066, No. U2241212.
This work is also supported by funding from Tencent Technology Co., Ltd (P0039546).





\bibliographystyle{elsarticle-num-names} 
\bibliography{main}




\end{document}